\documentclass[sigconf]{acmart}
\AtBeginDocument{%
  }
\usepackage{booktabs}
\usepackage{multirow}
\usepackage[table]{xcolor}
\usepackage{pifont}
\usepackage{enumitem}

\newcommand{\cmark}{\textcolor{green!60!black}{\ding{51}}}
\newcommand{\xmark}{\textcolor{red!70!black}{\ding{55}}}

\copyrightyear{2026}
\acmYear{2026}
\setcopyright{cc}
\setcctype{by}
\acmConference[WWW '26]{Proceedings of the ACM Web Conference 2026}{April 13--17, 2026}{Dubai, United Arab Emirates}
\acmBooktitle{Proceedings of the ACM Web Conference 2026 (WWW '26), April 13--17, 2026, Dubai, United Arab Emirates}
\acmPrice{}
\acmDOI{10.1145/3774904.3792674}
\acmISBN{979-8-4007-2307-0/2026/04}

\newcommand{\modelname}{\texttt{TaxoBell}}

\begin{document}

\title{ \modelname: Gaussian Box Embeddings for Self-Supervised Taxonomy Expansion}

\author{Sahil Mishra}
\affiliation{%
 \institution{Indian Institute of Technology Delhi}
 \city{New Delhi}
 \country{India}}
 \email{sahil.mishra@ee.iitd.ac.in}

\author{Srinitish Srinivasan}
\affiliation{%
 \institution{Indian Institute of Technology Delhi}
 \city{New Delhi}
 \country{India}}
 \email{nitish@ee.iitd.ac.in}

 \author{Srikanta Bedathur}
\affiliation{%
 \institution{Indian Institute of Technology Delhi}
 \city{New Delhi}
 \country{India}}
 \email{srikanta@iitd.ac.in}

\author{Tanmoy Chakraborty}
\affiliation{%
  \institution{Indian Institute of Technology Delhi}
  \city{New Delhi}
  \country{India}
}
\affiliation{%
  \city{Abu Dhabi}
  \country{UAE}
}
 \email{tanchak@iitd.ac.in}

\renewcommand{\shortauthors}{Sahil Mishra, Srinitish Srinivasan, Srikanta Bedathur, and Tanmoy Chakraborty}

\begin{abstract}
Taxonomies form the backbone of structured knowledge representation across diverse domains, enabling applications such as e-commerce and semantic search. Yet, manual taxonomy expansion is labor-intensive and slow. Existing methods rely on point-based vector embeddings, which model symmetric similarity and thus struggle with the asymmetric relationships that are fundamental to taxonomies. Box embeddings offer a promising alternative by enabling containment and disjointness, but they face key issues: (i) unstable gradients at the intersection boundaries, (ii) no notion of semantic uncertainty, and (iii) limited capacity to represent polysemy or ambiguity. We address these shortcomings with \modelname, a Gaussian box embedding framework that translates between box geometries and multivariate Gaussian distributions, where means encode semantic location and covariances encode uncertainty. Energy-based optimization yields stable optimization, robust modeling of ambiguous concepts, and interpretable hierarchical reasoning. Extensive experiments on five benchmark datasets demonstrate that \modelname\ significantly outperforms eight state-of-the-art taxonomy expansion baselines by 19\% in MRR and around 25\% in Recall@k. We further demonstrate the advantages and pitfalls of \modelname\ with error analysis and ablation studies.

\end{abstract}


\maketitle

\section{Introduction}
\label{sec:intro}

Taxonomies are domain-centric hierarchical structures that encode hypernymy (``is-a'') relations among concepts and entities, underpinning various applications. E-commerce platforms such as Amazon and Alibaba organize products to streamline navigation and retrieval \cite{mao2020octet,luo2020alicoco,karamanolakis2020txtract,zhang2014taxonomy}, while Pinterest curates taxonomies of \textit{home decor} and \textit{fashion styles} to improve visual search \cite{mahabal2023producing}. 

\begin{figure}[!t]
\centering
\includegraphics[width=0.99\columnwidth]{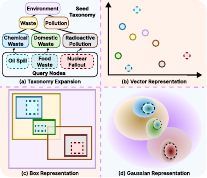}
\Description{A description of the intro.pdf image. Replace this text with an appropriate description.}
\caption{Overview of taxonomy expansion and \modelname.}
\Description{A diagram illustrating taxonomy expansion and the contribution of the proposed model.}
\label{fig:intro}
\end{figure}

Despite their utility, most real-world taxonomies are still built and maintained by domain experts. This manual process is slow, expensive, and increasingly complex to sustain as data grows and new concepts appear every day. In practice, coverage lags behind what applications need, which can harm search quality, recommendation accuracy, and user experience. To keep pace, taxonomies must evolve continuously and at scale. This motivates automated \emph{taxonomy expansion} in which, given an existing ``seed'' taxonomy, the goal is to add new entities, also called ``query nodes'', by placing each one under the most appropriate parent node. Automated expansion reduces human effort, shortens update cycles, and helps prevent taxonomies from becoming stale in fast-changing domains. An instance of taxonomy expansion is shown in Fig. \ref{fig:intro}(a), where query nodes ``\textit{Oil Spill}'', ``\textit{Food Waste}'' and ``\textit{Nuclear Fallout}'' are placed under the most appropriate parents, which are ``\textit{Chemical Waste}'', ``\textit{Domestic Waste}'' and ``\textit{Radioactive Pollution}''. For consistency, we refer to the child as the query and the parent as the anchor in accordance with prior work \cite{wang2021enquire,flame,shen2020taxoexpan}.


{\bf Summary of existing work on taxonomy expansion.} In taxonomy expansion, methods typically fall into two streams, namely \emph{semantic} and \emph{structural} methods. Early semantic approaches infer hypernymy from lexical patterns such as Hearst \cite{snow2004learning,hearst1992automatic,panchenko2016taxi} or distributional embeddings \cite{chang2017distributional, Mishra2025QuanTaxoAQ}. More recent works adopt self-supervision, harvesting \(\langle\)parent, child\(\rangle\) pairs from a seed taxonomy to train hypernymy classifiers; some methods, such as HiExpan \cite{shen_hiexpan_2018} and STEAM  \cite{yu_steam_2020}, rely on corpus-derived features, while others, such as leveraging external textual descriptions of surface forms. However, they are limited by insufficient training data and fall short of fully exploiting the taxonomy's inherent structural information. To incorporate structure, contemporary methods such as LORex \cite{mishra2025rank}, TEMP \cite{liu2021temp} and TaxoEnrich \cite{jiang2022taxoenrich} add signals from full taxonomy paths, STEAM \cite{yu_steam_2020} utilizes mini-paths, and TaxoExpan \cite{shen2020taxoexpan} and HEF \cite{wang2021enquire} use local ego-graphs. Despite these advances, a core limitation of these methods is that they embed entities in Euclidean space as vectors, which are agnostic to asymmetric relationships and hierarchy possessed by the taxonomic geometry as shown in Fig. \ref{fig:intro}(b). To better align representation with asymmetric structure, a parallel line of work models entities with \emph{box embeddings} \cite{jiang2023single}, where each concept is an axis-aligned hyperrectangle and hypernym pairs are expressed through \emph{containment}, meaning a child's region lies entirely within its parent's as shown in Fig. \ref{fig:intro}(c). 

{\bf Limitations of existing work.} While box embeddings capture asymmetric ``is-a'' structure via directional containment, which point (vector) embeddings cannot do, current formulations such as BoxTaxo \cite{jiang2023single} and TaxBox \cite{taxbox} remain limited for several practical reasons. First, their geometric training objectives defined over the intersection are typically piecewise and often suffer from vanishing or unstable gradients at disjoint boundaries, yielding weak or noisy learning signals for both centers and offsets. Secondly, boxes provide no principled way to represent \emph{interpretable uncertainty} as their boundaries are hard margins, so ambiguity, polysemy, and noisy evidence cannot be expressed as calibrated confidence over space. Therefore, the core challenge is to achieve stable, asymmetric hierarchy modeling and principled uncertainty representation within a unified framework.

{\bf Our contributions.} We model taxonomy entities as \textit{Gaussian boxes}, which are axis-aligned hyperrectangles equipped with a multivariate Gaussian density that captures semantic location using \emph{mean} ($\mu$) and concept generality using \emph{covariance} ($\Sigma$). Unlike mapping a point vector to a Gaussian distribution, which lacks hierarchical inductive bias and can fake inclusion by inflating or rotating covariance, Gaussian boxes occupy a geometric region and, unlike a hard box, assign calibrated probability mass within the region as shown in Fig. \ref{fig:intro}(d). This enables richer asymmetric relations such as \emph{probabilistic containment}, \emph{probabilistic disjointness}, and \emph{partial overlap}. Moreover, the desired level of uncertainty is naturally controllable via confidence intervals, where the $1\sigma$, $2\sigma$, and $3\sigma$ enclose roughly $68\%$, $95\%$, and $99.7\%$ of the probability mass, respectively, letting us set containment and overlap criteria at a chosen level and derive axis-wise box extents from $\Sigma$. We optimize the Gaussian boxes with energy-based objectives rather than probabilistic, geometric, or dot-product objectives on means, which fail to incorporate covariances \cite{vilnis2014word}. In contrast, energy functions score pairs of Gaussians by taking inner products between the distributions themselves (including covariance) and providing smooth, closed-form signals for Gaussians. Therefore, we define two energy functions: one for symmetric overlap to model semantic coherence and one for asymmetric overlap to learn hierarchical containment.

In this paper, we introduce \modelname\footnote{The name \modelname\ highlights the bell-shaped curve of the Gaussian distribution. Code: https://github.com/sahilmishra0012/TaxoBell.}, which translates between axis-aligned boxes and multivariate Gaussian parameterization and optimizes them with self-supervised, energy-based learning to capture semantic, asymmetric ``is-a'' relations, preserving calibrated uncertainty through the covariance, enabling robust placement of unseen entities for taxonomy expansion. 

Specifically, we make the following contributions.

\textbf{First}, we use self-supervision to derive training data directly from the seed taxonomy, without extra annotations. Each $\langle$child, parent$\rangle$ edge forms a positive instance, while negatives are sampled from non-ancestors in the child's neighborhood. This forces the model to separate the true parent from overly general ancestors and local confounders, yielding strong supervision from taxonomy structure at minimal cost.

\textbf{Secondly}, we project $\langle$child, parent$\rangle$ pairs to Gaussian boxes and jointly train them with two complementary energy functions, inspired by \cite{vilnis2014word}. A symmetric overlap Bhattacharyya distance term encourages high probabilistic overlap for true pairs, while an asymmetric KL term enforces directional containment by pulling the child distribution inside its parent. To avoid degenerate solutions, we add volume regularization that keeps covariances well-conditioned and prevents variance collapse. At inference time, candidate parents are ranked by these energy functions, and the chosen Gaussian can be translated to a box by fixing a confidence level to control retained uncertainty.

\textbf{Thirdly}, we conduct an extensive set of experiments on five real-world taxonomy benchmarks against eight state-of-the-art baselines. Results demonstrate that \modelname\ consistently outperforms vector and geometric representations, such as box embeddings, improving Mean Rank (MR) by ~\textbf{43\%}, Mean Reciprocal Rank (MRR) by ~\textbf{19\%}, Recall@k by ~\textbf{25\%}, and Hit@k by ~\textbf{21\%}. We further present ablations on the projection design and energy-based optimization, along with case studies showing Gaussian-box representations and efficient attachment to the seed taxonomy.

\section{Related Work}
\label{sec:related-work}
\noindent{\bf Taxonomy Expansion.}
Prior work expands taxonomies in two main ways. Corpus-based methods extract hypernyms from textual distance or contextual similarity \cite{shen_hiexpan_2018,huang_corel_2020,lee2022taxocom}, but require suitable corpora and often treat expansion as edge classification with limited global hierarchy awareness. Seed-taxonomy methods instead learn structure from existing paths \cite{jiang2022taxoenrich, yu_steam_2020, shen2020taxoexpan, wang2021enquire, liu2021temp, xutaxoprompt}: ego-network approaches focus on local context \cite{shen2020taxoexpan, wang2021enquire}, while path/walk models capture broader organization \cite{liu2021temp, xutaxoprompt}. Yet most represent nodes as points, which favors symmetric similarity and weakly encodes asymmetric hypernymy. We use Gaussian box embeddings to directly model directional containment, better aligning with taxonomy structure.

{\bf Box Representation.}
Box embeddings naturally encode asymmetric hierarchy better than point vectors \cite{patel2020representing,vilnis2018probabilistic,li2018smoothing}. Gumbel “soft-edge” boxes ease optimization but blur crisp, human-interpretable margins \cite{dasgupta2020improving}. Prior work shows geometry can improve interpretability \cite{chheda2021box,ren2020query2box}, and \citet{jiang2023single} blends these ideas for taxonomy modeling. However, intersection-based box losses remain piecewise and fragile near boundaries, with uncertainty only implicit. We maintain the box's inductive bias but introduce an explicit Gaussian density, yielding smooth, closed-form containment energies and calibrated per-dimension uncertainty while preserving the asymmetry crucial for taxonomy expansion.

{\bf Gaussian Representation.}
Gaussian embeddings model each concept as a distribution: the mean encodes its position, while the covariance captures uncertainty, generality, or polysemy. Early work showed that closed-form objectives such as expected likelihood and KL divergence let Gaussians represent confidence and asymmetric entailment \cite{vilnis2014word}. Later extensions introduced multi-sense/mixture models \cite{athiwaratkun2017multimodal,athiwaratkun-etal-2018-probabilistic} and applied Gaussians to knowledge graphs (KG2E) to handle confidence and one-to-many relations \cite{kg2e}. Probability product kernels connect Gaussian overlap to positive-definite similarity \cite{Jebara}, while KL naturally enforces directional containment. Energy-based training is smooth, jointly shaping means and covariances and avoiding brittle, piecewise losses \cite{lecun2006tutorial}. Building on these ideas, we map axis-aligned boxes to Gaussians and optimize paired energies for semantic similarity and asymmetric containment with calibrated uncertainty.

\section{Preliminaries}
\label{sec:problemform}

\subsection{Taxonomy Expansion} 

\begin{definition}
    \textbf{Taxonomy.} A taxonomy $\mathcal{T} = (\mathcal{N}, \mathcal{E})$ is defined as a directed acyclic graph where each node $n \in \mathcal{N}$ represents a concept with a surface name (a word or a phrase) while $\langle n_p,n_c \rangle \in \mathcal{E}$ represents an edge in the graph which represents a hypernymy relationship from parent $n_p$ to child $n_c$. 
\end{definition}

\begin{definition}
    \textbf{Taxonomy Expansion.} The taxonomy expansion task is to add a set of terms $\mathcal{C}$ to the given {\em seed taxonomy} $\mathcal{T}^0=(\mathcal{N}^0,\mathcal{E}^0)$ such that it returns an expanded taxonomy $\mathcal{T}=(\mathcal{N},\mathcal{E})=(\mathcal{N}^0\cup\mathcal{C},\mathcal{E}^0\cup\mathcal{R})$ where $\mathcal{R}$ represents newly added relations between seed nodes in $\mathcal{N}^0$ and new nodes in  $\mathcal{C}$. Specifically, for each query node  $n_q\in\mathcal{C}$, the model treats every seed node $a\in\mathcal{N}^0$ as an anchor, computes a matching score $f(a,n_q)$, and selects the best parent $n_p=\arg\max_{a\in\mathcal{N}^0} f(a,n_q)$. The edge $\langle n_p,n_q\rangle$ is then added to $\mathcal{R}$, expanding the seed taxonomy $\mathcal{T}^0$ to $\mathcal{T}$.
\end{definition}

\subsection{Box Representation}
A box is parameterized by a pair of $d$-dimensional vectors $b=(c,o)$, where the \emph{center} $c\in\mathbb{R}^d$ and the \emph{offsets} $o\in\mathbb{R}^{d^+}$ define the axis-aligned hyperrectangle $\prod_{i=1}^d [\,c_i - o_i,\; c_i + o_i\,]$ \cite{jiang2023single,ren2020query2box}. The positivity constraint $o_i>0$ in every dimension ensures a non-degenerate box. Its volume is $\operatorname{Vol}(b)\;=\;\prod_{i=1}^d (r_i-l_i),$ where $r_i=c_i+o_i$ and $l_i=c_i-o_i$.
Following \citet{jiang2023single}, we define three pairwise relationships between the boxes: (i) \textbf{Enclosure}: $b_x \cap b_y=b_x$, where $b_y$ encloses $b_x$; (ii) \textbf{Intersection}: $b_x \cap b_y = b_z \neq \emptyset$, where $b_z$ is the intersection box of $b_x$ and $b_y$ and has offset greater than 0; (iii) \textbf{Disjointness}: $b_x \cap b_y = b_z = \emptyset$, where $b_z$ is an imaginary intersection box of $b_x$ and $b_y$ and has offset less than or equal to 0. The intersection of the boxes $b_x$ and $b_y$ also leads to the computation of the conditional probability between these two boxes as, $P\left(b_y \mid b_x\right) = {P\left(b_x, b_y\right)}/{P\left(b_x\right)} = {\operatorname{Vol}\left(b_x \cap b_y\right)}/{\operatorname{Vol}\left(b_x\right)}$.

\subsection{Multivariate Gaussian Preliminaries}


\paragraph{Gaussian distribution.}
A multivariate Gaussian distribution $g = (\mu, \Sigma)$ can be described by its mean vector $\mu \in \mathbb{R}^d$ and symmetric positive definite covariance matrix $\Sigma  \in \mathbb{R}^{d \times d}$. A random variable $X \in \mathbb{R}^d$ is \emph{Gaussian} ( $X\sim\mathcal{N}(\mu,\Sigma)$) if its density $\mathcal{N}(x\mid\mu,\Sigma)$ is, 
\begin{equation}
    \begin{split}
    p(x)=\frac{1}{(2\pi)^{d/2}\sqrt{\det\Sigma}}\exp\!\Big[-\tfrac12(x-\mu)^\top\Sigma^{-1}(x-\mu)\Big],
    \end{split}
\end{equation}
where the mean $\mu$ specifies the \emph{location} of the variable in embedding space, and the covariance $\Sigma$ controls its \emph{uncertainty} along each direction, analogous to an offset vector. Intuitively,
for a Gaussian, about $68\%$, $95\%$, and $99.7\%$ of the mass lies within $1\sigma$, $2\sigma$, and $3\sigma$ of the mean (the “$68$–$95$–$99.7$ rule”). For computational stability we use axis-aligned Gaussians with $\Sigma=\mathrm{diag}(\sigma_1^2,\ldots,\sigma_d^2)\succ0$, so inverses and log-determinants are elementwise and $\mathcal{O}(d)$. 

\paragraph{Energy functions.}
An \emph{energy function} \cite{lecun2006tutorial, vilnis2014word} $E_\theta(x,y)$ assigns a scalar score to an input–output pair $(x,y)$, parameterized by $\theta$. The goal of energy-based learning is to tune $\theta$ so \emph{positive} pairs have lower energy than \emph{negatives} under a contrastive loss $\mathcal{L}$. For Gaussian embeddings, interactions between two distributions $P=\mathcal{N}(\mu_1,\Sigma_1)$ and $Q=\mathcal{N}(\mu_2,\Sigma_2)$ are generally captured by two kinds of complementary energy functions,
\begin{itemize}[leftmargin=*]
    \item A \emph{symmetric similarity function}, which computes the dot product between the means of $P$ and $Q$ but does not incorporate the covariances and therefore discards probabilistic information. A more principled choice for a symmetric energy function is the inner product of the distributions themselves as $k(P,Q)\;=\langle P,Q\rangle \;=\; \int_{\mathbb{R}^d} P(x)\,Q(x)\,dx$ which is the $\alpha{=}1$ case of the \emph{probability product kernel} (PPK) family \cite{Jebara}, $k_{\alpha}(P,Q)=\int_{\mathbb{R}^d}P(x)^{\alpha}Q(x)^{\alpha}dx$ for $\alpha>0$. Setting $\alpha=\tfrac12$ yields the \emph{Bhattacharyya coefficient},
    \begin{equation}
    \mathrm{BC}(P,Q)\;=\;k_{1/2}(P,Q)\;=\;\int_{\mathbb{R}^d}\!\sqrt{P(x)\,Q(x)}\,dx\;\in\;(0,1],
    \end{equation}
    and its {Bhattacharyya distance} $        D_B(P,Q) = -\log \mathrm{BC}(P,Q).$
    Both $k(P, Q)$ and $\mathrm{BC}(P,Q)$ increase with greater symmetric overlap, i.e., when the means are closer and the covariances are more compatible. $\mathrm{BC}$ has the closed form with,
    \begin{equation}
    D_B(P,Q)=\tfrac18(\mu_1-\mu_2)^\top \Sigma_m^{-1}(\mu_1-\mu_2)+\tfrac12\log\frac{\det\Sigma_m}{\sqrt{\det\Sigma_1\,\det\Sigma_2}},
    \end{equation}
    where $\Sigma_m=\tfrac12(\Sigma_1+\Sigma_2)$. Hence, for Gaussian densities $P$ and $Q$, we have $\mathrm{BC}(P,Q)=\exp\!\big(-D_B(P,Q)\big)$, providing a smooth, probabilistically grounded symmetric similarity.
    
    \item An \emph{asymmetric} energy, which computes directionality instead of just similarity, is the Kullback–Leibler (KL) divergence from $P$ to $Q$, measuring the information lost when approximating $P$ with $Q$ as $D_{\mathrm{KL}}(P\!\parallel\!Q)\;=\;\int_{\mathbb{R}^d} P(x)\, \log\frac{P(x)}{Q(x)}\,dx$.
    For Gaussian densities $P$ and $Q$, this admits the closed form,
    \begin{equation}
    \begin{split}
    D_{\mathrm{KL}}(P\!\parallel\!Q)= \tfrac12\Big[\operatorname{tr}(\Sigma_2^{-1}\Sigma_1&)+(\mu_2-\mu_1)^\top \Sigma_2^{-1}(\mu_2-\mu_1) \\&-d + \log\frac{\det\Sigma_2}{\det\Sigma_1}\Big].
    \end{split}
    \end{equation}
    Low $D_{\mathrm{KL}}(P\!\parallel\!Q)$ means that $P$ is (softly) contained in $Q$, while large values signal poor coverage.
\end{itemize}

These energies act directly on $\mu$ and $\Sigma$, yield smooth gradients, and, under diagonal covariances, are inexpensive to compute, making them well suited for training Gaussian embeddings with both similarity (overlap) and hierarchy (containment) signals.
\begin{figure}[!t]
    \centering
    \includegraphics[width=0.99\linewidth]{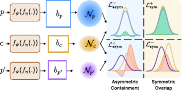}
    \caption{Overview of \modelname. Entities are encoded with $f_\eta(.)$, mapped to axis-aligned boxes using $f_\psi(.)$, and then projected to Gaussian embeddings. Training optimizes two energies on the Gaussians -- a symmetric overlap term (BC) and an asymmetric containment term (KL).}
    \Description{Overview diagram of the \modelname\ framework, showing entities encoded, mapped to boxes, and projected to Gaussian embeddings.}
    \label{fig:model}
\end{figure}

\section{The \modelname\ Framework}
\label{sec:model}

\subsection{Overview}
In \modelname\ (Fig.~\ref{fig:model}), entities are encoded from text into axis-aligned boxes (Sec.~\ref{subsec:boxproject}) and then mapped to multivariate Gaussians (Sec.~\ref{subsec:gaussproject}). Using the seed taxonomy for self-supervision, we train these Gaussians with two energies that (i) maximize symmetric parent--child overlap and (ii) enforce asymmetric hierarchical containment (Sec.~\ref{subsec:train}).

\subsection{Box Projection}
\label{subsec:boxproject}
Taxonomy nodes are conceptual entities, which are generally represented by surface names and definitions in natural language. To obtain a semantic vector for each entity $e$, we use a pre–trained language model $f_\eta(.)$ (BERT \cite{devlin_bert_2019} in our case). We construct the input sequence by concatenating the surface name and its definition with the BERT delimiter tokens ``[\texttt{CLS}]'' and ``[\texttt{SEP}]'' as, $x= [\texttt{CLS}] \; e \; [\texttt{SEP}] \; s \; [\texttt{SEP}]$ where $s$ denotes the definition of the entity with surface name $e$. The encoder produces contextual token representations of ``[\texttt{CLS}]'', ``[\texttt{SEP}]'', $e$ and $s$. Therefore, we take the final‐layer \texttt{[CLS]} embedding as the representation of node $e$ as, $\mathbf{n} \;=\; f_\eta(x)[0] \in \mathbb{R}^{k}$, where index 0 represents the position of embedding of [\texttt{CLS}] token while $k$ is the embedding size.

Next, we project the entity representation $n$ to a box $b=(c,o)$, where center $\mathbf{c}\in\mathbb{R}$ and offset $\mathbf{o}\in\mathbb{R}^{d^+}$ are vector embeddings with dimension $d$ encoder $f_\psi(.)$.  Concretely, we use two independent 2-layered multilayer perceptrons (MLPs), namely $\mathrm{MLP}_c$ and $\mathrm{MLP}_o$, to map to the center and offset, respectively, as,
\begin{align}
{c} = \text{MLP}_{c}({n})
&= {W}^{(2)}_{c}\,\sigma\big({W}^{(1)}_{c}{n}+{b}^{(1)}_{c}\big)+{b}^{(2)}_{c},
\label{eq:center}\\
{o} = \phi(\text{MLP}_{o}({n}))
&= \phi({W}^{(2)}_{o}\,\sigma\big({W}^{(1)}_{o}{n}+{b}^{(1)}_{o}\big)+{b}^{(2)}_{o}),
\label{eq:offset}
\end{align}
where $\sigma(\cdot)$ is a pointwise nonlinearity (e.g., ReLU or GELU) and $W^{i}_c$ and $W^{i}_o$ are weights corresponding to layers of MLPs of center and offset projectors, respectively. To ensure a valid box with strictly positive offsets in every dimension, we apply a monotone positivity function $\phi$, such as exponential or softplus, to the offset. This gives us boxes $b_p=(c_p,o_p)$, $b_{p'}=(c_{p'},o_{p'})$ and $b_c=(c_c,o_c)$ for parent, negative parent and child respectively.

\subsection{Gaussian Projection}
\label{subsec:gaussproject}
To enable boxes to model calibrated uncertainty, we project them to a multivariate Gaussian distribution $g = (\mu, \Sigma)$, where $\mu$ is the mean, while $\Sigma$ is the covariance matrix. Intuitively, the box center $c$ already locates the concept in space, so we identify it with the Gaussian mean, $\boldsymbol{\mu}=\mathbf{c}$. The offset governs dispersion and thus induces axis-aligned uncertainty. Concretely, we construct a positive semi-definite covariance by squaring the offsets and placing them on the diagonal as $\Sigma=\operatorname{diag}(\mathbf{o}\odot\mathbf{o}),$ $\Sigma_{ij}=o_i^2\,\delta_{ij}.$

In particular, we obtain three Gaussian distributions,  $\mathcal{N}_p({\mu}_p,{\Sigma}_p)$, $\mathcal{N}_{p'}({\mu}_{p'},{\Sigma}_{p'})$ and $\mathcal{N}_c({\mu}_c,{\Sigma}_c)$, for parent, negative parent and child respectively. This ``box-to-Gaussian'' projection of the parent preserves the geometric role of offsets, where a large offset means a wider box face and, correspondingly, a larger marginal variance. The induced equal-density contours of $\mathcal{N}(\boldsymbol{\mu},\boldsymbol{\Sigma})$ are axis-aligned hyper-ellipsoids nested around $\boldsymbol{\mu}$; their sizes at 1/2/3 standard deviations provide an interpretable notion of semantic scope (e.g., the 68–95–99.7 rule), which a hard box cannot express. Moreover, this diagonal assumption is computationally efficient while still capturing concept generality through anisotropic variances. It also creates a tight link between the box ``volume'' and Gaussian spread: $\det(\boldsymbol{\Sigma})=\prod_{i=1}^d o_i^2$, so $\tfrac{1}{2}\log\det\boldsymbol{\Sigma}=\sum_{i=1}^d \log o_i$ can be used as a differentiable proxy for semantic volume to prevent the problem of vanishing gradients. Crucially, moving from hard boxes to Gaussians replaces brittle, piecewise geometry with smooth overlap, enabling stable optimization with probabilistic objectives. Probabilistic containment of a child inside a parent is no longer a binary event but a graded relation governed jointly by mean separation and relative covariance, which our losses can shape continuously.

\subsection{Gaussian Box Training and Inference}
\label{subsec:train}
\textbf{Self-Supervised Generation.} We now optimize the Gaussian box embeddings so that they faithfully encode taxonomic hierarchy, i.e., asymmetric child–parent relations, under a self-supervised paradigm. The seed taxonomy $\mathcal{T}^0=(\mathcal{N}^0,\mathcal{E}^0)$ provides natural supervision where every $\langle\text{child},\text{anchor}\rangle$ edge is treated as a positive pair for training, where child node $n_c$ is the ``query'' and parent node $n_p$ is the ``anchor''. To shape fine-grained decision boundaries, we mine $N$ \emph{hard negatives} ($\{n_{p'_i}|_{i=1}^{N}\subset\mathcal{N}^0\}$) from the local neighborhood of the query node $n_c$, which are its siblings, uncles, cousins and grandparents. These negatives are topologically close to the child and semantically confusable with the parent, thus providing informative contrast to create distinctive boundaries. These $N{+}1$ pairs (one positive and $N$ negatives) constitute a single training instance $\mathbf{X}_{\langle n_p,n_c\rangle}=\big\{\langle n_c,n_p,n_{p'_1}\rangle,\langle n_c,n_p,n_{p'_2}\rangle,\dots,\langle n_c,n_p,n_{p'_N}\rangle\big\}$. Repeating this procedure for every edge in $\mathcal{E}^0$ yields the full self-supervision set $\mathbf{\mathcal{X}}=\{\mathbf{X}_{\langle n_p,n_c\rangle}\mid \langle n_p,n_c\rangle\in\mathcal{E}^0\}$. During training, each triple $\langle n_c,n_p,n_{p'_i}\rangle$ is encoded and then projected into a box and a Gaussian, which is optimized using energy-based optimization so that the positive Gaussian distributions overlap.

\noindent\textbf{Energy-Based Optimization.} We optimize the Gaussian boxes end-to-end by combining two complementary criteria: \textit{symmetric overlap}, which captures conceptual similarity, and an \textit{asymmetric} term that encodes hierarchical directionality. In addition, we apply regularization to prevent collapse to a point mass and to keep covariances well-conditioned. The overall objective aggregates the following components.

\noindent\paragraph{1. Symmetric Overlap.} To capture direction-agnostic semantic similarity between a child $n_c$ and its parent $n_p$, we use the Bhattacharyya coefficient $BC(\mathcal{N}_p, \mathcal{N}_c) = \exp\left(-D_B(\mathcal{N}_p, \mathcal{N}_c)\right)$. $\mathrm{BC}$ is high when means are close and covariances are compatible, and it decreases as centers separate or uncertainty profiles diverge. The symmetric overlap loss for a triple $\langle n_c,n_p,n_{p'_i}\rangle$ is computed using binary cross entropy as,
\begin{equation}
\mathcal{L}_{\text{sym}} = - \log\left(BC(\mathcal{N}_p, \mathcal{N}_c)\right) - \log\left(1 - BC(\mathcal{N}_{p'}, \mathcal{N}_c)\right)
\label{eq:bce_loss_canonical}
\end{equation}

\begin{table*}[!t]
\centering
\caption{Performance comparison of \modelname\ with baseline methods. Each entry in the table reports $\text{mean}^\text{std-dev}$ in percentage over five independent runs with distinct random seeds. The best performance is marked in bold, while the best baseline is underlined. For MR (not a percentage), lower values indicate better performance and are marked with ``$\downarrow$''.}
\label{table:main_results}

\setlength{\tabcolsep}{4pt}
\scalebox{0.9}{
\begin{tabular}{@{}l|rrrrr|rrrrr|rrrrr@{}}
\toprule
\multirow{2}{*}{\shortstack[l]{\textbf{Methods}}}
& \multicolumn{5}{c|}{\textbf{Science}} & \multicolumn{5}{c|}{\textbf{Environment}} & \multicolumn{5}{c}{\textbf{WordNet}} \\
\cmidrule(lr){2-6}\cmidrule(lr){7-11}\cmidrule(lr){12-16}
& \textbf{MR$\downarrow$} & \textbf{MRR} & \textbf{R@1} & \textbf{R@5} & \textbf{Wu\&P}
& \textbf{MR$\downarrow$} & \textbf{MRR} & \textbf{R@1} & \textbf{R@5} & \textbf{Wu\&P}
& \textbf{MR$\downarrow$} & \textbf{MRR} & \textbf{R@1} & \textbf{R@5} & \textbf{Wu\&P} \\
\midrule
BERT+MLP    & $\text{241.6}^{\text{23.1}}$ & $\text{21.3}^{\text{3.7}}$ & $\text{13.1}^{\text{4.1}}$ & $\text{27.3}^{\text{3.3}}$ & $\text{45.7}^{\text{1.7}}$
             & $\text{74.2}^{\text{7.2}}$ & $\text{21.4}^{\text{2.6}}$ & $\text{11.1}^{\text{3.0}}$ & $\text{31.8}^{\text{2.1}}$ & $\text{48.2}^{\text{0.4}}$
             & $\text{314.6}^{\text{62.2}}$ & $\text{16.8}^{\text{2.1}}$ & $\text{9.1}^{\text{2.5}}$ & $\text{38.1}^{\text{3.7}}$ & $\text{41.4}^{\text{1.2}}$ \\
TaxoExpan    & $\text{117.6}^{\text{15.7}}$ & $\text{40.1}^{\text{2.8}}$ & $\text{24.6}^{\text{3.9}}$ & $\text{41.8}^{\text{3.1}}$ & $\text{55.7}^{\text{1.2}}$
             & $\text{56.1}^{\text{5.3}}$ & $\text{28.4}^{\text{3.1}}$ & $\text{12.9}^{\text{5.7}}$ & $\text{37.1}^{\text{2.3}}$ & $\text{49.8}^{\text{1.0}}$
             & $\text{141.6}^{\text{17.3}}$ & $\text{31.1}^{\text{2.2}}$ & $\text{17.1}^{\text{3.5}}$ & $\text{38.1}^{\text{1.6}}$ & $\text{49.7}^{\text{1.8}}$ \\
Arborist      & $\text{83.1}^{\text{7.2}}$ & $\text{41.2}^{\text{3.1}}$ & $\text{26.5}^{\text{4.4}}$ & $\text{51.5}^{\text{3.5}}$ & $\text{61.2}^{\text{1.4}}$
             & $\text{39.1}^{\text{5.8}}$ & $\text{33.7}^{\text{4.7}}$ & $\text{24.9}^{\text{5.9}}$ & $\text{45.2}^{\text{2.1}}$ & $\text{52.5}^{\text{1.3}}$             
             & $\text{92.6}^{\text{3.6}}$ & $\text{33.7}^{\text{2.7}}$ & $\text{20.3}^{\text{3.5}}$ & $\text{43.9}^{\text{2.8}}$ & $\text{53.5}^{\text{1.3}}$ \\
BoxTaxo      & $\text{67.7}^{\text{11.3}}$ & $\text{43.0}^{\text{3.8}}$ & $\text{25.3}^{\text{4.5}}$ & $\text{49.2}^{\text{3.1}}$ & $\text{63.1}^{\text{1.7}}$
             & $\text{34.1}^{\text{7.3}}$ & $\text{41.6}^{\text{4.9}}$ & $\text{32.3}^{\text{6.2}}$ & $\text{51.4}^{\text{3.5}}$ & $\text{65.1}^{\text{1.4}}$             
             & $\text{77.1}^{\text{8.8}}$ & $\text{34.1}^{\text{3.2}}$ & $\text{22.3}^{\text{4.2}}$ & $\text{45.7}^{\text{3.6}}$ & $\text{58.1}^{\text{1.7}}$ \\
TMN          & $\text{54.2}^{\text{5.1}}$ & $\text{45.5}^{\text{2.5}}$ & $\text{31.5}^{\text{3.8}}$ & $\text{53.7}^{\text{1.9}}$ & $\text{65.7}^{\text{1.2}}$
             & $\text{31.3}^{\text{3.5}}$ & $\text{43.8}^{\text{2.1}}$ & $\text{34.7}^{\text{3.7}}$ & $\text{51.1}^{\text{3.2}}$ & $\text{63.9}^{\text{2.0}}$
             & $\text{73.9}^{\text{4.9}}$ & $\text{35.8}^{\text{2.7}}$ & $\text{23.7}^{\text{3.2}}$ & $\text{49.0}^{\text{2.6}}$ & $\text{56.6}^{\text{0.8}}$ \\
STEAM        & $\text{31.7}^{\text{3.3}}$ & $\text{50.7}^{\text{3.5}}$ & $\text{34.8}^{\text{4.9}}$ & $\text{59.1}^{\text{4.1}}$ & $\text{72.2}^{\text{1.3}}$
             & $\text{27.1}^{\text{2.8}}$ & $\text{44.2}^{\text{2.7}}$ & $\text{34.1}^{\text{3.4}}$ & $\text{55.6}^{\text{2.9}}$ & $\text{65.2}^{\text{1.7}}$
             & $\text{61.1}^{\text{2.4}}$ & $\text{37.3}^{\text{2.1}}$ & $\text{24.9}^{\text{3.9}}$ & $\text{54.5}^{\text{1.7}}$ & $\text{59.2}^{\text{1.2}}$ \\
TaxoEnrich       & $\text{22.1}^{\text{3.8}}$ & $\text{55.2}^{\text{4.9}}$ & $\text{39.5}^{\text{5.1}}$ & $\text{65.4}^{\text{3.1}}$ & $\text{72.7}^{\text{2.2}}$
             & $\text{17.2}^{\text{6.1}}$ & $\text{49.8}^{\text{4.3}}$ & $\text{42.1}^{\text{5.3}}$ & $\text{61.0}^{\text{1.2}}$ & $\text{71.8}^{\text{2.5}}$             
             & $\text{54.3}^{\text{3.5}}$ & $\text{45.1}^{\text{3.1}}$ & $\text{27.2}^{\text{4.3}}$ & $\text{61.7}^{\text{4.3}}$ & $\text{68.7}^{\text{1.9}}$ \\
\cmidrule{1-16}
\textbf{\modelname$_{BC}$}
  & ${\textbf{13.2}}^{\textbf{0.9}}$ & $\underline{\text{58.2}}^{\text{1.2}}$ & $\textbf{49.0}^{\textbf{1.9}}$ & $\textbf{74.8}^{\textbf{1.3}}$ & $\textbf{76.2}^{\textbf{0.4}}$
  & $\textbf{8.3}^{\textbf{0.5}}$  & $\textbf{58.7}^{\textbf{2.5}}$ & $\underline{\text{46.5}}^{\text{4.4}}$ & $\underline{\text{75.0}}^{\text{1.4}}$ & $\textbf{77.2}^{\textbf{1.0}}$
  & $\textbf{40.9}^{\textbf{4.2}}$ & $\underline{\text{50.3}}^{\text{1.4}}$ & $\underline{\text{36.1}}^{\text{0.8}}$ & $\underline{\text{70.3}}^{\text{2.4}}$ & $\underline{\text{73.4}}^{\text{0.3}}$ \\
\textbf{\modelname$_{KL}$}
  & $\underline{\text{13.8}}^{\text{1.2}}$ & $\textbf{58.5}^{\textbf{1.6}}$ & $\underline{\text{48.0}}^{\text{2.8}}$ & $\underline{\text{74.1}}^{\text{0.6}}$ & $\underline{\text{75.6}}^{\text{0.9}}$
  & $\underline{\text{8.9}}^{\text{0.7}}$  & $\underline{\text{58.4}}^{\text{1.7}}$ & $\textbf{47.3}^\textbf{3.8}$ & $\textbf{75.0}^{\textbf{1.3}}$ & $\underline{\text{75.8}}^{\text{0.5}}$
  & $\textbf{41.0}^{\textbf{4.2}}$ & $\textbf{51.2}^{\textbf{1.2}}$ & $\textbf{37.1}^{\textbf{0.6}}$ & $\textbf{70.4}^{\textbf{2.1}}$ & $\textbf{73.5}^{\textbf{0.8}}$ \\

\bottomrule
\end{tabular}
}
\setlength{\tabcolsep}{7.57pt}
\scalebox{0.9}{

\begin{tabular}{@{}l|rrrrrr|rrrrrr@{}}
\toprule
\textbf{Methods} &
\multicolumn{6}{c|}{\textbf{SemEval-Food}} &
\multicolumn{6}{c}{\textbf{MeSH}} \\
\cmidrule(lr){2-7}\cmidrule(lr){8-13}
& \textbf{MR$\downarrow$} & \textbf{MRR} & \textbf{R@1} & \textbf{R@5} & \textbf{H@1} & \textbf{H@5}
& \textbf{MR$\downarrow$} & \textbf{MRR} & \textbf{R@1} & \textbf{R@5} & \textbf{H@1} & \textbf{H@5} \\
\midrule
BERT+MLP
& $\text{508.8}^{\text{35.3}}$ & $\text{47.1}^{\text{2.1}}$ & $\text{12.3}^{\text{2.7}}$ & $\text{26.1}^{\text{2.9}}$ & $\text{13.8}^{\text{1.2}}$ & $\text{25.7}^{\text{0.2}}$
& $\text{1352.4}^{\text{57.1}}$ & $\text{9.3}^{\text{1.7}}$ & $\text{1.1}^{\text{0.6}}$ & $\text{5.2}^{\text{0.9}}$ & $\text{3.7}^{\text{0.3}}$ & $\text{7.2}^{\text{0.3}}$ \\
TaxoExpan
& $\text{343.8}^{\text{17.2}}$ & $\text{41.3}^{\text{1.7}}$ & $\text{25.1}^{\text{2.2}}$ & $\text{32.8}^{\text{2.3}}$ & $\text{14.2}^{\text{0.7}}$ & $\text{27.3}^{\text{0.9}}$
& $\text{891.1}^{\text{31.4}}$ & $\text{17.1}^{\text{2.1}}$ & $\text{2.9}^{\text{0.8}}$ & $\text{10.3}^{\text{0.9}}$ & $\text{7.3}^{\text{0.5}}$ & $\text{15.8}^{\text{0.1}}$ \\
BoxTaxo
& $\text{363.2}^{\text{12.8}}$ & $\text{45.2}^{\text{1.6}}$ & $\text{27.3}^{\text{2.3}}$ & $\text{36.4}^{\text{2.1}}$ & $\text{29.1}^{\text{1.8}}$ & $\text{41.1}^{\text{0.5}}$
& $\text{620.2}^{\text{22.9}}$ & $\text{21.5}^{\text{2.9}}$ & $\text{17.1}^{\text{1.1}}$ & $\text{30.2}^{\text{0.6}}$ & $\text{16.5}^{\text{1.2}}$ & $\text{31.7}^{\text{1.1}}$ \\
Arborist
& $\text{247.9}^{\text{21.1}}$ & $\text{45.3}^{\text{1.9}}$ & $\text{29.3}^{\text{2.1}}$ & $\text{37.7}^{\text{2.2}}$ & $\text{21.5}^{\text{2.7}}$ & $\text{37.2}^{\text{1.4}}$
& $\text{553.6}^{\text{26.2}}$ & $\text{21.3}^{\text{2.3}}$ & $\text{19.0}^{\text{1.7}}$ & $\text{31.1}^{\text{1.3}}$ & $\text{16.7}^{\text{1.4}}$ & $\text{29.4}^{\text{0.9}}$ \\
TMN
& $\text{192.6}^{\text{16.1}}$ & $\text{51.8}^{\text{2.1}}$ & $\text{34.1}^{\text{2.6}}$ & $\text{43.2}^{\text{1.9}}$ & $\text{25.3}^{\text{2.1}}$ & $\text{40.9}^{\text{1.1}}$
& $\text{433.7}^{\text{16.5}}$ & $\text{23.5}^{\text{1.2}}$ & $\text{18.1}^{\text{0.6}}$ & $\text{33.2}^{\text{0.9}}$ & $\text{16.6}^{\text{1.8}}$ & $\text{31.8}^{\text{1.0}}$ \\
STEAM
& $\text{155.9}^{\text{14.0}}$ & $\text{53.8}^{\text{2.5}}$ & $\text{39.1}^{\text{3.1}}$ & $\text{51.3}^{\text{1.7}}$ & $\text{34.4}^{\text{2.3}}$ & $\text{58.7}^{\text{0.4}}$
& $\text{372.6}^{\text{16.2}}$ & $\text{25.1}^{\text{3.1}}$ & $\text{17.7}^{\text{2.7}}$ & $\text{35.1}^{\text{1.0}}$ & $\text{18.2}^{\text{1.8}}$ & $\text{38.2}^{\text{1.5}}$ \\
TaxoEnrich
& $\text{101.7}^{\text{11.2}}$ & $\text{53.7}^{\text{3.3}}$ & $\text{41.8}^{\text{2.7}}$ & $\text{57.5}^{\text{1.1}}$ & $\text{39.2}^{\text{1.9}}$ & $\text{67.2}^{\text{1.2}}$
& $\text{247.7}^{\text{15.5}}$ & $\text{25.3}^{\text{2.7}}$ & $\text{18.7}^{\text{2.1}}$ & $\text{37.6}^{\text{1.3}}$ & $\text{21.3}^{\text{1.9}}$ & $\text{47.1}^{\text{1.2}}$ \\
\cmidrule{1-13}
\textbf{\modelname$_{BC}$}
& $\textbf{36.3}^{\textbf{7.3}}$ & $\textbf{58.1}^{\textbf{1.7}}$ & $\textbf{46.07}^{\textbf{2.1}}$ & $\underline{\text{76.0}}^{\text{0.4}}$ & $\textbf{45.1}^{\textbf{1.7}}$ & $\underline{\text{75.2}}^{\text{0.8}}$
& $\textbf{175.4}^{\textbf{4.7}}$ & $\underline{\text{28.9}}^{\text{1.9}}$ & $\textbf{20.2}^{\textbf{2.8}}$ & $\underline{\text{42.0}}^{\text{1.1}}$ & $\underline{\text{24.5}}^{\text{1.7}}$ & $\underline{\text{54.3}}^{\text{1.1}}$ \\
\textbf{\modelname$_{KL}$}
& $\underline{\text{36.6}}^{\text{6.7}}$ & $\underline{\text{57.4}}^{\text{2.0}}$ & $\underline{\text{44.79}}^{\text{3.4}}$ & $\textbf{76.1}^{\textbf{1.6}}$ & $\underline{\text{45.1}}^{\text{2.5}}$ & $\textbf{75.6}^{\textbf{1.1}}$
& $\underline{\text{175.4}}^{\text{9.1}}$ & $\textbf{29.2}^{\textbf{2.2}}$ & $\underline{\text{20.1}}^{\text{3.1}}$ & $\textbf{42.3}^{\textbf{1.5}}$ & $\textbf{24.8}^{\textbf{2.4}}$ & $\textbf{55.3}^{\textbf{1.9}}$ \\
\bottomrule
\end{tabular}
}

\end{table*}

%
%

\begin{table}[t]
\centering
\caption{Combined Statistical Test Results using Fisher's Method comparing \modelname\ to the best baseline.}
\label{table:statistics}

\centering
\begin{tabular}{@{}l|ccc@{}}
\toprule
& \textbf{SCI} & \textbf{ENV} & \textbf{WordNet} \\
\midrule
\(\boldsymbol{\chi^2}\) 
& 90.71 & 73.01 & 100.62 \\
\textbf{\(p\)-value} 
& \(1.34\times 10^{-15}\) & \(7.56\times 10^{-12}\) & \(9.61\times 10^{-18}\) \\
\bottomrule
\end{tabular}
\begin{tabular}{@{}l|cc@{}}
& \textbf{SemEval-Food} & \textbf{MeSH} \\
\midrule
\(\boldsymbol{\chi^2}\) 
& 85.66 & 62.38 \\
\textbf{\(p\)-value} 
& \(8.53\times 10^{-13}\) & \(1.58\times 10^{-8}\) \\
\bottomrule
\end{tabular}

\end{table}

\begin{table}[t]
\centering
\caption{Impact of the asymmetric optimization ($\mathcal{L}_\text{asym}$) on SCI, MeSH, and Food benchmarks. ``$\downarrow$'' indicates that lower values denote better performance. `W/O' means \textit{without}.}
\label{table:no_kl}

\setlength{\tabcolsep}{1pt}
\scalebox{0.85}{
\begin{tabular}{@{}c|rrr|rrr|rrr@{}}
\toprule
\addlinespace[2pt]
\multirow{2}{*}{\textbf{Method}} &
\multicolumn{3}{c|}{\textbf{Science}} &
\multicolumn{3}{c|}{\textbf{Food}} &
\multicolumn{3}{c}{\textbf{MeSH}} \\
\cmidrule(lr){2-4}\cmidrule(lr){5-7}\cmidrule(lr){8-10}
& \textbf{H@1} & \textbf{MR$\downarrow$} & \textbf{MRR}
& \textbf{H@1} & \textbf{MR$\downarrow$} & \textbf{MRR}
& \textbf{H@1} & \textbf{MR$\downarrow$} & \textbf{MRR} \\
\midrule
\textbf{W/O $\mathcal{L}_\text{asym}$} & 42.35 & 13.44 & 55.07 & 33.89 & 49.39 & 48.51 & 22.70 & 249.32 & 25.86 \\
\rowcolor{gray!20}\textbf{$\mathcal{L}_\text{asym}$} & 51.70 & 12.77 & 58.50 & 46.93 & 36.60 & 59.44 & 26.50 & 175.35 & 29.15 \\
\midrule
$\uparrow$\% & +22.08 & +4.99 & +6.23 & +38.48 & +25.90 & +22.53 & +16.74 & +29.67 & +12.72 \\
\bottomrule
\end{tabular}}

\end{table}

\begin{table}[t]
\centering
\caption{Impact of the symmetric optimization ($\mathcal{L}_\text{sym}$) on SCI, MeSH, and Food benchmarks. ``$\downarrow$'' indicates that lower values denote better performance. `W/O' means \textit{without}.}
\label{table:no_bc}

\setlength{\tabcolsep}{1pt}
\scalebox{0.90}{
\begin{tabular}{@{}c|rrr|rrr|rrr@{}}
\toprule
\addlinespace[2pt]
\multirow{2}{*}{\textbf{Method}} &
\multicolumn{3}{c|}{\textbf{Science}} &
\multicolumn{3}{c|}{\textbf{Food}} &
\multicolumn{3}{c}{\textbf{MeSH}} \\
\cmidrule(lr){2-4}\cmidrule(lr){5-7}\cmidrule(lr){8-10}
& \textbf{H@1} & \textbf{MR$\downarrow$} & \textbf{MRR}
& \textbf{H@1} & \textbf{MR$\downarrow$} & \textbf{MRR}
& \textbf{H@1} & \textbf{MR$\downarrow$} & \textbf{MRR} \\
\midrule
\textbf{W/O $\mathcal{L}_\text{sym}$} & 11.77 & 196.57 & \,5.75 & 25.77 & \,75.67 & 39.26 & 11.24 & 322.75 & 18.89 \\
\rowcolor{gray!20}\textbf{$\mathcal{L}_\text{sym}$}    & 50.91 & \,14.09 & 59.41 & 43.43 & \,29.94 & 56.35 & 23.18 & 171.76 & 27.99 \\
\midrule
$\uparrow\%$   & +332.5 & +92.8 & +933.2 & +68.5 & +60.4 & +43.5 & +106.3 & +46.8 & +48.2 \\
\bottomrule
\end{tabular}}

\end{table}

\noindent\paragraph{2. Asymmetric Overlap.} To encode hierarchical directionality, we align the child distribution to its parent using the KL divergence $D_{\mathrm{KL}}(\mathcal{N}_c\!\parallel\!\mathcal{N}_p)$, which quantifies the information lost when $\mathcal{N}_p$ approximates $\mathcal{N}_c$. If the child's mass is well contained within the parent, the value of $D_{\mathrm{KL}}(\mathcal{N}_c\!\parallel\!\mathcal{N}_p)$ is small compared to negative pairs. We contrast this against a hard negative parent $\mathcal{N}_{p'}$ with a margin-based triplet objective:
\begin{equation}
    \mathcal{L}_{\text{align}} = \max\left(0, D_{KL}(\mathcal{N}_c \| \mathcal{N}_p) - D_{KL}(\mathcal{N}_c \| \mathcal{N}_{p'}) + \delta\right),
    \label{eq:align_loss}
\end{equation}
where $\delta>0$ enforces a minimum separation between the positive and negative pairs.

However, just enforcing $\mathcal{L}_{\text{align}}$ does not control the coverage of the parent because a narrowly peaked parent may still yield a small KL, also reducing the child to a small peak. To ensure parents remain broader than their children in order to accommodate more children, we introduce a reverse-KL term that is coupled to the log-volume gap, $D_{KL}(\mathcal{N}_p \| \mathcal{N}_c) \ge \text{logVol}(\mathcal{N}_p) - \text{logVol}(\mathcal{N}_c)=D_{rep}$. We penalize violations of the coverage constraint with a hinge, as
\begin{equation}
\mathcal{L}_{\text{diverge}}
=\max\!\left\{0,\; C\times D_{{rep}} -
D_{{KL}}(\mathcal{N}_p \parallel \mathcal{N}_c) \right\},
\label{eq:repel_loss}
\end{equation}
where $C>0$ scales the required informational separation by the geometric
(volume) separation. The final asymmetric objective is a weighted combination,
\begin{equation}
\mathcal{L}_{\text{asym}}
=\mathcal{L}_{\text{align}} + \lambda\,\mathcal{L}_{\text{diverge}},
\label{eq:total_loss}
\end{equation}
with hyperparameter $\lambda>0$, yielding stable training and consistent hierarchical alignment. We discuss the geometrical implications of eq.\ref{eq:repel_loss} and \ref{eq:total_loss} for different values of $C$ and $\lambda$ and their impact on model performance in Appendix \ref{app:scale}. 

\noindent\paragraph{3. Volume Regularization}
To prevent numerical instability and avoid degenerate Gaussians, we introduce two regularizers. The first regularizer, \textit{minimum volume regularization}, prevents volume collapse by enforcing a minimum standard deviation. The minimum regularization loss for an entity distribution $\mathcal{N}_x(\mu_x,\Sigma_x)$ is formulated as a squared hinge loss,
\begin{equation}
    \mathcal{L}_{\text{reg}}(\mathcal{N}_x)=\frac{1}{d}\,\big\|\,(\delta_{\mathrm{var}} I-\Sigma_x)_{+}\,\big\|_{F}^{2},
\end{equation}
where $\delta_{\mathrm{var}}>0$ is the minimum variance, $I$ is an identity matrix of dimension $d$, $(\cdot)_+$ denotes the element-wise hinge such that $(A_{ij})_+=\max(0,A_{ij})$ while $\|\cdot\|_{F}$ is the Frobenius norm.

The second regularizer, \textit{covariance clipping}, enhances stability by preventing variances from becoming excessively large, which can lead to floating-point errors. This term penalizes variances that exceed a maximum threshold, \(M_{\mathrm{var}}\),
\begin{equation}
    \mathcal{L}_{\text{clip}}(\mathcal{N}_x)
= \frac{1}{d}\,\operatorname{tr}\!\big( [\,\Sigma_x - M_{\mathrm{var}} I\,]_+ \big),
\end{equation}
where \([A]_+\) denotes the elementwise hinge \(\max(0,A_{ij})\), \(\operatorname{tr}(\cdot)\) is the trace, \(I\) is the identity, \(d\) is the embedding dimension, and \(M_{\mathrm{var}}>0\) is the variance ceiling. These volume regularizations are computed for parent, child, and negative parent distributions ($\mathcal{N}_p, \mathcal{N}_c, \mathcal{N}_{p'}$).

Therefore, we combine the above objective functions as follows,
\begin{equation}
    \mathcal{L}_{\text{overall}}=\mathcal{L}_{\text{sym}}+\mathcal{L}_{\text{asym}}+\mathcal{L}_{\text{reg}}+\mathcal{L}_{\text{clip}}
    \label{eq:combined_loss_function}
\end{equation}

\noindent\textbf{Inference.} 
During inference, the goal is to attach a new query concept to an appropriate parent (anchor) in the seed taxonomy. Given a query $q$, we encode it with the pretrained encoder, project it to a box $b_q=(c_q,o_q)$, and obtain its Gaussian embedding $\mathcal{N}_q(\mu_q,\Sigma_q)$ (Sec. \ref{subsec:boxproject}, \ref{subsec:gaussproject}). For each candidate anchor $a$ in the seed taxonomy, with Gaussian $\mathcal{N}_a(\mu_a,\Sigma_a)$, we compute two scores -- a \emph{symmetric overlap} term via the Bhattacharyya coefficient $\mathrm{BC}(\mathcal{N}_a,\mathcal{N}_q)$, and an \emph{asymmetric containment} term via the KL $D_{\mathrm{KL}}(\mathcal{N}_q\!\parallel\!\mathcal{N}_a)$ separately, naming two models \modelname$_{BC}$ and \modelname$_{KL}$. For visualization or downstream rules, the chosen Gaussian can be translated back to a box by fixing a confidence level (e.g., $1\sigma$, $2\sigma$) and mapping per-dimension spreads to offsets, preserving the uncertainty.

\noindent\textbf{Gaussian to Box Translation.} 
To render a multivariate Gaussian $\mathcal{N}(\mu,\Sigma)$ as an axis-aligned box $b=(c,o)$, we set the center to the mean $c=\mu$ and the offset to $o = k\,\sqrt{\Sigma}$. Choosing $k\in\{1,2,3\}$ gives roughly 68\%, 95\%, and 99.7\% coverage per dimension, respectively.

\section{Experiments}
We evaluate \modelname\ on five real-world taxonomies against vector- and geometry-based baselines using standard ranking/retrieval metrics. The implementation details are in Appendix~\ref{app:implementation}.

\subsection{Experimental Setup}
\label{subsec:experimentsetup}
\subsubsection{Datasets}
\label{subsubsec:data}

Following \cite{jiang2022taxoenrich,liu2021temp,xu-etal-2023-tacoprompt}, we evaluate \modelname\ on five public taxonomies spanning general and specialized domains for comprehensive coverage across major domains -- (i) Environment (ENV), (ii) Science (SCI) from SemEval-2016 Task 13 \cite{bordea-etal-2016-semeval}, (iii) Food from SemEval-2015 Task 17 \cite{bordea2015semeval}, (iv) WordNet sub-taxonomies from  \citet{bansal-etal-2014-structured}, and (v) Medical Subject Headings (MeSH), a widely used clinical taxonomy \cite{lipscomb2000medical}, where (i), (ii) and (iv) are single parent taxonomies while others are multi-parent ones. Full dataset statistics and train-test split are discussed in Appendix~\ref{app:data}.

\subsubsection{Baselines}
\label{subsubsec:baseline}

We aim to study the effectiveness of Gaussian box embeddings over vector and geometric embeddings for representing and expanding taxonomic hierarchies. We evaluate \modelname\ against seven baselines spanning vector and structure-aware approaches: (i) \textbf{BERT+MLP} \cite{panchenko2016taxi},  (ii) \textbf{Arborist} \cite{manzoor2020expanding}, (iii) \textbf{TaxoExpan} \cite{shen2020taxoexpan}, (iv) \textbf{TMN} \cite{zhang2021taxonomy}, (v) \textbf{STEAM} \cite{yu_steam_2020}, (vi) \textbf{BoxTaxo} \cite{jiang2023single}, and (vii) \textbf{TaxoEnrich} \cite{jiang2022taxoenrich}. The baselines are discussed in Appendix \ref{app:baselines}.
  
\subsubsection{Evaluation Metrics}
\label{subsubsec:eval}

Following prior works \cite{wang_qen_2022,xu-etal-2023-tacoprompt,Mishra2025QuanTaxoAQ}, for each query node $q$, we rank all candidate anchors in the seed taxonomy and record the rank(s) of the gold parent(s). We report \textbf{Mean Rank (MR)}, \textbf{Mean Reciprocal Rank (MRR)}, \textbf{Recall@$k$}, \textbf{Hit@$\boldsymbol{k}$} and \textbf{Wu \& Palmer}. Metric definitions and the choices for single and multi-parent taxonomies are detailed in Appendix~\ref{app:metrics}.

\begin{table}[t]
\centering
\caption{Effect of the volume regularization ($\mathcal{L}_\text{reg}$ and $\mathcal{L}_\text{clip}$) on SCI, MeSH, and Food benchmarks. ``$\downarrow$'' indicates that lower values denote better performance. `W/O' means \textit{without}.}
\label{table:vol_reg}

\setlength{\tabcolsep}{1pt}
\scalebox{0.90}{
\begin{tabular}{@{}c|rrr|rrr|rrr@{}}
\toprule
\addlinespace[2pt]
\multirow{2}{*}{\textbf{Method}} &
\multicolumn{3}{c|}{\textbf{Science}} &
\multicolumn{3}{c|}{\textbf{Food}} &
\multicolumn{3}{c}{\textbf{MeSH}} \\
\cmidrule(lr){2-4}\cmidrule(lr){5-7}\cmidrule(lr){8-10}
& \textbf{H@1} & \textbf{MR$\downarrow$} & \textbf{MRR}
& \textbf{H@1} & \textbf{MR$\downarrow$} & \textbf{MRR}
& \textbf{H@1} & \textbf{MR$\downarrow$} & \textbf{MRR} \\
\midrule
\textbf{W/O Reg} & 43.52 & 20.14 & 50.34 & 39.83 & 47.55 & 46.05 & 22.92 & 198.61 & 26.16 \\
\rowcolor{gray!20}\textbf{Reg} & 51.70 & 12.77 & 58.50 & 46.93 & 36.60 & 59.44 & 26.50 & 175.35 & 29.15 \\
\midrule
\textbf{$\uparrow\%$} & +18.80 & +36.59 & +16.21 & +17.83 & +23.03 & +29.08 & +15.62 & +11.71 & +11.43 \\
\bottomrule
\end{tabular}}

\end{table}

\begin{table}[t]
\centering
\caption{Effect of $\mathcal{L}_\text{diverge}$ on SCI, MeSH, and Food benchmarks. ``$\downarrow$'' indicates that lower values denote better performance. `W/O' means \textit{without}.}
\label{table:diverge}

\setlength{\tabcolsep}{1pt}
\scalebox{0.85}{
\begin{tabular}{@{}c|rrr|rrr|rrr@{}}
\toprule
\addlinespace[2pt]
\multirow{2}{*}{\textbf{Method}} &
\multicolumn{3}{c|}{\textbf{Science}} &
\multicolumn{3}{c|}{\textbf{Food}} &
\multicolumn{3}{c}{\textbf{MeSH}} \\
\cmidrule(lr){2-4}\cmidrule(lr){5-7}\cmidrule(lr){8-10}
& \textbf{H@1} & \textbf{MR$\downarrow$} & \textbf{MRR}
& \textbf{H@1} & \textbf{MR$\downarrow$} & \textbf{MRR}
& \textbf{H@1} & \textbf{MR$\downarrow$} & \textbf{MRR} \\
\midrule
\textbf{W/O $\mathcal{L}_\text{diverge}$} & 45.88 & 24.81 & 55.86 & 38.93 & 43.77 & 50.25 & 23.02 & 214.14 & 26.21 \\
\rowcolor{gray!20}\textbf{$\mathcal{L}_\text{diverge}$} & 51.70 & 12.77 & 58.50 & 46.93 & 36.60 & 59.44 & 26.50 & 175.35 & 29.15 \\
\midrule
\textbf{$\uparrow\%$} & +12.69 & +48.53 & +4.73 & +20.55 & +16.38 & +18.29 & +15.12 & +18.11 & +11.22 \\
\bottomrule
\end{tabular}}

\end{table}

%
%

\subsection{Comparative Analysis and Statistical Tests}
Table \ref{table:main_results} presents the results of all baselines compared with two variants of our method -- \modelname$_{KL}$, with negative KL divergence as ranker, and \modelname$_{BC}$ with Bhattacharyya coefficient as ranker. Vector embedding methods, such as BERT+MLP, perform the worst, indicating that symmetric similarity in a point space is ill-suited to hierarchical attachment, while path-based methods such as TaxoExpan, Arborist, STEAM, and TaxoEnrich improve over point baselines but still lag behind our Gaussian-box models. Both \modelname$_{KL}$ and \modelname$_{BC}$ consistently achieve the lowest MR and the highest MRR/Recall across datasets, with the largest gains on multi-parent taxonomies. Between our variants, \modelname$_{KL}$ strongest R@$1$, reflecting sharper asymmetric containment, while \modelname$_{BC}$ often matches or slightly improves MR/MRR, suggesting smoother overlap. In sum, representing concepts as Gaussian boxes, retaining box geometry while modeling per-dimension uncertainty, yields state-of-the-art ranking quality on both single- and multi-parent benchmarks.

\noindent\textbf{Statistical tests.} We compare \modelname\ to the strongest baseline (TaxoEnrich) and, per dataset, combine metric-level p-values using Fisher's method, which under the null follows a $\chi^2$ distribution with 2$k$ degrees of freedom, where $k$ is the number of metrics. Table \ref{table:statistics} reports both $X^2$ and the combined $p$-value, showing large $\chi^2$ values and small $p$-values on every dataset, so we reject the null hypothesis that \modelname\ and the best baseline perform equivalently and conclude that \modelname's gains are statistically significant.

\begin{figure}
    \centering
    \includegraphics[width=1.0\linewidth]{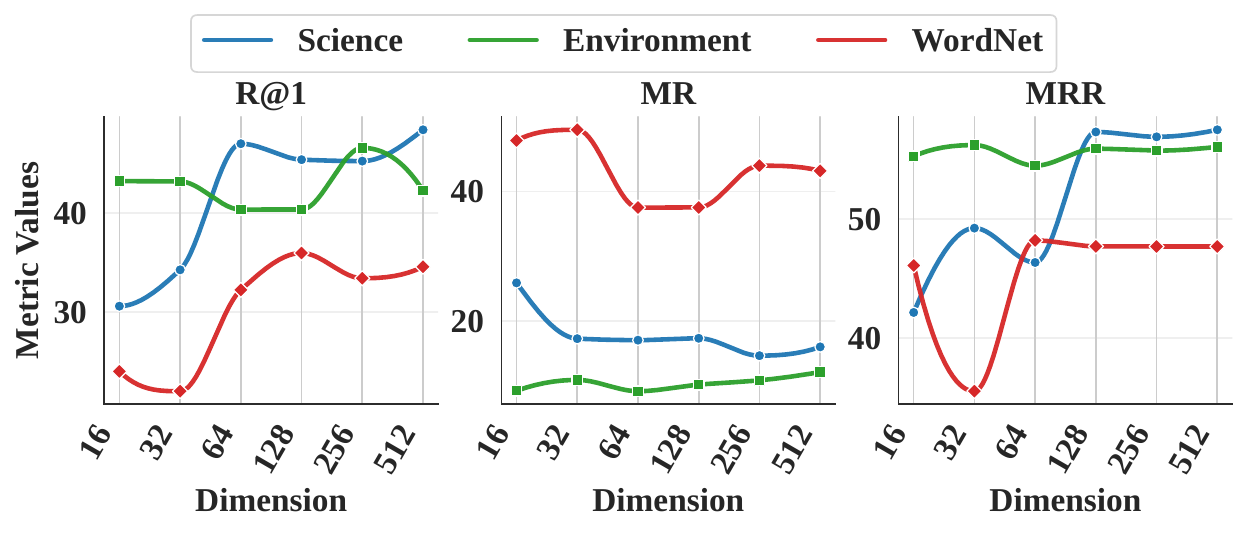}
    \caption{Effect of embedding dimensionality on performance for SCI, ENV, and WordNet benchmarks.}
    \Description{Plot showing the effect of embedding dimensionality on performance for SCI, ENV, and WordNet benchmarks.}
    \label{fig:dimensions}
    
\end{figure}

\subsection{Ablation Studies}
\label{subsec:ablations}
We ablate \modelname\ on SCI, Food, and MeSH to isolate component effects and compare alternatives. Additional ablations and hyperparameter studies are in Appendices~\ref{app:ablations} and \ref{app:scale}, respectively.

\subsubsection{Impact of Asymmetric and Symmetric Losses}
We study the effect of both complementary losses, $\mathcal{L}_{\text{asym}}$ and $\mathcal{L}_{\text{sym}}$, on the performance of \modelname\ as shown in Tables \ref{table:no_kl} and \ref{table:no_bc}. Removing $\mathcal{L}_{\text{asym}}$ erases the notion of directionality, so queries cluster near related anchors but attach to structurally wrong nodes such as siblings, uncles, or overly general ancestors, yielding higher MR and lower R@1/MRR. Furthermore, dropping $\mathcal{L}_{\text{sym}}$ leaves the model only with the KL optimization, which can be satisfied by shrinking the child or inflating the parent, degrading semantic geometry and calibration -- again hurting MR and R@1/MRR. Using both losses jointly gives calibrated, non-degenerate embeddings and the strongest ranks, showing hierarchy and similarity are complementary signals for reliable attachment. 

\subsubsection{Impact of Volume Regularization}
We also study the effect of volume regularization, consisting of two components, the minimum‐variance regularizer $\mathcal{L}_{\text{reg}}$ and the variance clipping term $\mathcal{L}_{\text{clip}}$, on model performance as shown in Table \ref{table:vol_reg}. We observe that removing these terms consistently hurts performance across datasets and metrics. Without a lower bound on variance, the model can shrink a child's covariance to satisfy KL, making parent and child nearly indistinguishable and destroying fine-grained ordering. Without an upper bound, it can inflate a parent to fake the containment, yielding overly broad ellipsoids that break hierarchical margins and destabilize training. Enforcing both bounds keeps covariances well-conditioned, yields calibrated uncertainty, preserves the intended relation, and produces more reliable containment decisions.

\begin{table}[t]
\centering
\caption{Direct vs. Gaussian projection variants of \modelname\ on SCI and ENV benchmarks. $D_{*}$ is vector embedding to Gaussian projection while $G_{*}$  is box-to-Gaussian projection. For MR, lower values indicate better performance.}
\label{table:direct_gauss}

\setlength{\tabcolsep}{4pt}
\begin{tabular}{@{}c|rrr|rrr@{}}
\toprule
\multirow{2}{*}{\textbf{Methods}} &
\multicolumn{3}{c|}{\textbf{Science}} &
\multicolumn{3}{c}{\textbf{Environment}} \\
\cmidrule(lr){2-4}\cmidrule(lr){5-7}
& \textbf{MR$\downarrow$} & \textbf{MRR} & \textbf{R@1}
& \textbf{MR$\downarrow$} & \textbf{MRR} & \textbf{R@1} \\
\midrule
\textbf{\modelname$_{D_{KL}}$} & 58.10 & 18.83 & \,5.82 & 44.62 & 25.50 & 19.23 \\
\rowcolor{gray!20}\textbf{\modelname$_{G_{KL}}$} & 13.96 & 58.50 & 51.76 & \,8.98 & 58.36 & 47.31 \\
\midrule
\textbf{$\uparrow\%$} & +76.0 & +210.7 & +789.3 & +79.9 & +128.9 & +146.0 \\
\midrule
\textbf{\modelname$_{D_{BC}}$} & 57.76 & 18.95 & 11.76 & 44.67 & 24.48 & 19.23 \\
\rowcolor{gray!20}\textbf{\modelname$_{G_{BC}}$} & 13.18 & 58.22 & 48.94 & \,8.31 & 58.68 & 46.54 \\
\midrule
\textbf{$\uparrow\%$} & +77.2 & +207.2 & +316.2 & +81.4 & +139.7 & +142.0 \\
\bottomrule
\end{tabular}

\end{table}

\begin{figure}
    \centering
\includegraphics[width=1.0\linewidth]{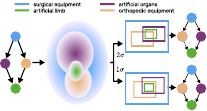}
    \caption{Case study of reconstructing a MeSH multi-parent subtree from learned Gaussians by controlling the uncertainty while converting Gaussians into boxes.}
    \Description{A visual representation of a MeSH multi-parent subtree reconstructed from learned Gaussians, illustrating the effect of uncertainty control.}
    \label{fig:case_study}
\end{figure}

\subsubsection{Impact of Reverse KL Divergence}
We study the impact of reverse-KL loss, $\mathcal{L}_{\text{diverge}}$ on \modelname's performance in Table \ref{table:diverge}. Using only the forward KL alignment lets the model minimize containment by making the parent narrowly peaked (or shrinking the child), producing brittle, poorly calibrated embeddings. Adding reverse KL imposes a coverage margin, stabilizes covariances, and improves separation among siblings. Empirically, this consistently lowers MR and raises H@1/MRR, acting as an uncertainty-margin regularizer that complements forward KL to better respect both hierarchy and semantics.

\subsubsection{Direct Projection vs Gaussian Projection}
We further study the role of geometry in the projection as shown in Table \ref{table:direct_gauss}. In a \emph{direct} variant, we map the encoder output $f_\psi(\cdot)$ straight to a Gaussian by using the hidden vector as $\mu$ and deriving a diagonal $\Sigma$ from its magnitudes. This performs worse on all benchmarks because uncertainty is tied to the encoder's feature basis, so the model cannot easily widen parents or separate siblings along the most useful directions. Our \modelname\  projection instead maps each concept to an axis-aligned box first and then converts that box into a Gaussian. This keeps a clear geometric bias (containment), provides interpretable variances, and lets training directly control spread to satisfy energy-based losses. The result is calibrated, non-degenerate Gaussians that capture asymmetric parent–child relations more faithfully and yield stronger rankings.

\subsection{Dimensionality Analysis}
To study how embedding size affects taxonomy expansion, we vary the dimension of the Gaussian boxes and plot the metrics in Fig.~\ref{fig:dimensions}. We observe that varying the embedding size reveals the expected capacity trade-off rather than a single monotone trend. Specifically on SCI, increasing the dimensions generally helps once the model has enough degrees of freedom to separate siblings and encode containment. ENV is more volatile as gains at smaller dimensions give way to a dip and only partial recovery at larger sizes, indicating sensitivity to over-parameterization relative to the data scale. WordNet saturates quickly, and moving from very small to moderate dimensions brings most of the improvement, after which the curves flatten. Overall, very small dimensions underfit, very large dimensions can overfit or destabilize covariance learning, and a mid-range dimensionality provides the best accuracy–stability balance.

\begin{table}[t]
\centering
\caption{Qualitative examples of \modelname\ on the Science and WordNet benchmarks. For each benchmark, we show one correct and one incorrect attachment. Correct predictions are marked with \cmark while incorrect with \xmark.}
\label{table:examples}
\setlength{\tabcolsep}{6pt}
\scalebox{0.85}{
\begin{tabular}{@{}lll@{}}
\toprule
\textbf{Query} & \textbf{Anchor (Gold)} & \textbf{Top Predictions} \\
\midrule
\rowcolor{gray!25}\multicolumn{3}{@{}c}{\textbf{Science}}\\
Solid Geometry & Geometry &
  \begin{tabular}[t]{@{}l@{}}
  Geometry \cmark\\
  Spherical geometry \xmark\\
  Plane geometry \xmark
  \end{tabular} \\
\rowcolor{gray!10}Orthopedics& Medical Science &
  \begin{tabular}[t]{@{}l@{}}
  Rheumatology\xmark\\
  Medicine \xmark\\
  Traumatology \xmark 
  \end{tabular} \\
\midrule
\rowcolor{gray!25}\multicolumn{3}{@{}c}{\textbf{WordNet}}\\
Convection & Temperature change &
  \begin{tabular}[t]{@{}l@{}}
  Temperature change \cmark\\
  Climate change \xmark\\
  Advection \xmark
  \end{tabular} \\
\rowcolor{gray!10}Infanticide & Murderer  &
  \begin{tabular}[t]{@{}l@{}}
  low-birth-weight baby \xmark\\
 matricide \xmark \\
   kittee \xmark 
  \end{tabular} \\
\bottomrule
\end{tabular}}

\end{table}

\subsection{Case Study and Error Analysis}
\label{subsec:case}
Fig. \ref{fig:case_study} illustrates an example of converting the MeSH subtree's Gaussians into boxes by choosing the confidence level. Using $1\sigma$, we get tighter boxes with lower uncertainty, cleanly capturing the intended multi-parent links, while with $2\sigma$, the boxes grow in size, increasing overlap and the child region can extend beyond the true shared intersection of its parents, which leads to confusion whether the child is multi-parent or misclassified edge due to erroneous intersection (shown by dashed arrow connecting \textit{orthopedic equipment} to \textit{artificial limb}). Furthermore, Table \ref{table:examples} complements this with single-parent examples from Science and WordNet. Typical successes include \textit{Solid geometry} placed under \textit{Geometry} in SCI because the definition and usage strongly match, which yields high symmetric overlap and \textit{Convection} placed under \textit{Temperature change} in WordNet since the query emphasizes heat transfer and temperature variation, so its mass sits well inside the anchor while remaining far from distractors like \textit{Climate change} or \textit{Heating}. We also observe two representative errors. In Science, \textit{Orthopedics} is attached to \textit{Rheumatology} instead of the gold \textit{Medical Science}, reflecting sibling ambiguity when several nearby anchors are semantically close. In WordNet, \textit{Infanticide} drifts toward unrelated or overly broad anchors, indicating weak domain grounding. Further case studies on multi-parent taxonomies are discussed in Appendix \ref{app:case_study}.

\section{Conclusion}
In this paper, we introduce \modelname, a self-supervised framework for taxonomy expansion that represents concepts as Gaussian boxes and optimizes them using two complementary energies, namely a symmetric Bhattacharyya overlap and an asymmetric KL-based hierarchical containment, resulting in a stable training recipe with volume regularization, which yields calibrated, non-degenerate embeddings that respect hierarchy while preserving semantic proximity. Across multiple taxonomy expansion benchmarks (single- and multi-parent), \modelname\ consistently reduces MR and improves MRR/Recall over strong vector, path-based, and hard-box baselines, and ablations verify that both losses, the learned Gaussian projection, and volume control are necessary for the gains. The representation is interpretable, probabilistic, and readily extensible. We believe probabilistic geometric representations of this kind offer a practical foundation for robust, scalable hierarchical reasoning in real-world systems.

\balance
\bibliographystyle{ACM-Reference-Format}
\bibliography{acm}

@inproceedings{liu2021temp,
	title        = {TEMP: taxonomy expansion with dynamic margin loss through taxonomy-paths},
	author       = {Liu, Zichen and Xu, Hongyuan and Wen, Yanlong and Jiang, Ning and Wu, Haiying and Yuan, Xiaojie},
	year         = 2021,
	booktitle    = {Proceedings of EMNLP},
	pages        = {3854--3863}
}

@inproceedings{xutaxoprompt,
	title        = {TaxoPrompt: A Prompt-based Generation Method with Taxonomic Context for Self-Supervised Taxonomy Expansion},
	author       = {Xu, Hongyuan and Chen, Yunong and Liu, Zichen and Wen, Yanlong and Yuan, Xiaojie},
	year         = 2022,
	month        = 7,
	booktitle    = {Proceedings of IJCAI},
	pages        = {4432--4438},
	note         = {Main Track},
}

@inproceedings{ren2020query2box,
	title        = {Query2box: Reasoning over Knowledge Graphs in Vector Space Using Box Embeddings},
	author       = {Hongyu Ren and Weihua Hu and Jure Leskovec},
	year         = 2020,
	booktitle    = {ICLR},
}

@inproceedings{li2018smoothing,
	title        = {Smoothing the Geometry of Probabilistic Box Embeddings},
	author       = {Xiang Li and Luke Vilnis and Dongxu Zhang and Michael Boratko and Andrew McCallum},
	year         = 2019,
	booktitle    = {ICLR},
}

@inproceedings{vilnis2018probabilistic,
	title        = {Probabilistic Embedding of Knowledge Graphs with Box Lattice Measures},
	author       = {Vilnis, Luke  and Li, Xiang  and Murty, Shikhar  and McCallum, Andrew},
	year         = 2018,
	booktitle    = {Proceedings of ACL},
	pages        = {263--272},
}

@inproceedings{bordea-etal-2016-semeval,
	title        = {{S}em{E}val-2016 Task 13: Taxonomy Extraction Evaluation ({TE}x{E}val-2)},
	author       = {Bordea, Georgeta  and Lefever, Els  and Buitelaar, Paul},
	year         = 2016,
	booktitle    = {Proceedings of SemEval},
	pages        = {1081--1091},
}

@inproceedings{shen2020taxoexpan,
	title        = {TaxoExpan: Self-supervised taxonomy expansion with position-enhanced graph neural network},
	author       = {Shen, Jiaming and Shen, Zhihong and Xiong, Chenyan and Wang, Chi and Wang, Kuansan and Han, Jiawei},
	year         = 2020,
	booktitle    = {Proceedings of WWW},
	pages        = {486--497}
}

@inproceedings{chang2017distributional,
	title        = {Distributional Inclusion Vector Embedding for Unsupervised Hypernymy Detection},
	author       = {Chang, Haw-Shiuan  and Wang, Ziyun  and Vilnis, Luke  and McCallum, Andrew},
	year         = 2018,
	booktitle    = {Proceedings of NAACL},
	pages        = {485--495},
}

@inproceedings{huang_corel_2020,
	title        = {Corel: Seed-guided topical taxonomy construction by concept learning and relation transferring},
	author       = {Huang, Jiaxin and Xie, Yiqing and Meng, Yu and Zhang, Yunyi and Han, Jiawei},
	year         = 2020,
	booktitle    = {Proceedings of KDD},
	pages        = {1928--1936}
}

@inproceedings{wang_qen_2022,
	title        = {Qen: Applicable taxonomy completion via evaluating full taxonomic relations},
	author       = {Wang, Suyuchen and Zhao, Ruihui and Zheng, Yefeng and Liu, Bang},
	year         = 2022,
	booktitle    = {Proceedings of WWW},
	pages        = {1008--1017}
}

@inproceedings{manzoor2020expanding,
	title        = {Expanding taxonomies with implicit edge semantics},
	author       = {Manzoor, Emaad and Li, Rui and Shrouty, Dhananjay and Leskovec, Jure},
	year         = 2020,
	booktitle    = {Proceedings of WWW},
	pages        = {2044--2054}
}

@inproceedings{hearst1992automatic,
	title        = {Automatic Acquisition of Hyponyms from Large Text Corpora},
	author       = {Hearst, Marti A.},
	year         = 1992,
	booktitle    = {{COLING} 1992 Volume 2: The 14th {I}nternational {C}onference on {C}omputational {L}inguistics},
}

@inproceedings{lee2022taxocom,
	title        = {Taxocom: Topic taxonomy completion with hierarchical discovery of novel topic clusters},
	author       = {Lee, Dongha and Shen, Jiaming and Kang, SeongKu and Yoon, Susik and Han, Jiawei and Yu, Hwanjo},
	year         = 2022,
	booktitle    = {Proceedings of WWW},
	pages        = {2819--2829}
}

@inproceedings{patel2020representing,
	title        = {Representing Joint Hierarchies with Box Embeddings},
	author       = {Dhruvesh Patel and Shib Sankar Dasgupta and Michael Boratko and Xiang Li and Luke Vilnis and Andrew McCallum},
	year         = 2020,
	booktitle    = {Automated Knowledge Base Construction},
}

@article{flame,
author = {Mishra, Sahil and Sudev, Ujjwal and Chakraborty, Tanmoy},
title = {FLAME: Self-Supervised Low-Resource Taxonomy Expansion Using Large Language Models},
year = {2024},
publisher = {Association for Computing Machinery},
address = {New York, NY, USA},
issn = {2157-6904},
journal = {ACM Trans. Intell. Syst. Technol.},
month = dec
}

@article{Mishra2025QuanTaxoAQ,
  title={QuanTaxo: A Quantum Approach to Self-Supervised Taxonomy Expansion},
  author={Mishra, Sahil and Patni, Avi and Chatterjee, Niladri and Chakraborty, Tanmoy},
  journal={arXiv preprint arXiv:2501.14011},
  year={2025},
  url={}
}

@inproceedings{wang2021enquire,
	title        = {Enquire one’s parent and child before decision: Fully exploit hierarchical structure for self-supervised taxonomy expansion},
	author       = {Wang, Suyuchen and Zhao, Ruihui and Chen, Xi and Zheng, Yefeng and Liu, Bang},
	year         = 2021,
	booktitle    = {Proceedings of WWW},
	pages        = {3291--3304}
}

@inproceedings{jiang2022taxoenrich,
	title        = {Taxoenrich: Self-supervised taxonomy completion via structure-semantic representations},
	author       = {Jiang, Minhao and Song, Xiangchen and Zhang, Jieyu and Han, Jiawei},
	year         = 2022,
	booktitle    = {Proceedings of WWW},
	pages        = {925--934}
}

@inproceedings{yu_steam_2020,
	title        = {Steam: Self-supervised taxonomy expansion with mini-paths},
	author       = {Yu, Yue and Li, Yinghao and Shen, Jiaming and Feng, Hao and Sun, Jimeng and Zhang, Chao},
	year         = 2020,
	booktitle    = {Proceedings of KDD},
	pages        = {1026--1035}
}

@inproceedings{devlin_bert_2019,
	title        = {{BERT}: Pre-training of Deep Bidirectional Transformers for Language Understanding},
	author       = {Devlin, Jacob  and Chang, Ming-Wei  and Lee, Kenton  and Toutanova, Kristina},
	year         = 2019,
	booktitle    = {Proceedings of NAACL},
	pages        = {4171--4186},
}

@inproceedings{shen_hiexpan_2018,
	title        = {Hiexpan: Task-guided taxonomy construction by hierarchical tree expansion},
	author       = {Shen, Jiaming and Wu, Zeqiu and Lei, Dongming and Zhang, Chao and Ren, Xiang and Vanni, Michelle T and Sadler, Brian M and Han, Jiawei},
	year         = 2018,
	booktitle    = {Proceedings of KDD},
	pages        = {2180--2189}
}

@inproceedings{chheda2021box,
	title        = {Box Embeddings: An open-source library for representation learning using geometric structures},
	author       = {Chheda, Tejas  and Goyal, Purujit  and Tran, Trang  and Patel, Dhruvesh  and Boratko, Michael  and Dasgupta, Shib Sankar  and McCallum, Andrew},
	year         = 2021,
	booktitle    = {Proceedings of EMNLP},
	pages        = {203--211},
}

@inproceedings{dasgupta2020improving,
	title        = {Improving Local Identifiability in Probabilistic Box Embeddings},
	author       = {Dasgupta, Shib and Boratko, Michael and Zhang, Dongxu and Vilnis, Luke and Li, Xiang and McCallum, Andrew},
	year         = 2020,
	booktitle    = {NeurIPS},
	volume       = 33,
	pages        = {182--192},
}

@inproceedings{vedula_enriching_2018,
	title        = {Enriching taxonomies with functional domain knowledge},
	author       = {Vedula, Nikhita and Nicholson, Patrick K and Ajwani, Deepak and Dutta, Sourav and Sala, Alessandra and Parthasarathy, Srinivasan},
	year         = 2018,
	booktitle    = {SIGIR},
	pages        = {745--754}
}

@inproceedings{zhang2021taxonomy,
	title        = {Taxonomy completion via triplet matching network},
	author       = {Zhang, Jieyu and Song, Xiangchen and Zeng, Ying and Chen, Jiaze and Shen, Jiaming and Mao, Yuning and Li, Lei},
	year         = 2021,
	booktitle    = {Proceedings of AAAI},
	volume       = 35,
	number       = 5,
	pages        = {4662--4670}
}

@inproceedings{karamanolakis2020txtract,
	title        = {{TX}tract: Taxonomy-Aware Knowledge Extraction for Thousands of Product Categories},
	author       = {Karamanolakis, Giannis  and Ma, Jun  and Dong, Xin Luna},
	year         = 2020,
	booktitle    = {Proceedings of ACL},
	pages        = {8489--8502},
}

@inproceedings{zhang2014taxonomy,
	title        = {Taxonomy discovery for personalized recommendation},
	author       = {Zhang, Yuchen and Ahmed, Amr and Josifovski, Vanja and Smola, Alexander},
	year         = 2014,
	booktitle    = {Proceedings of WSDM},
	pages        = {243--252},
}

@inproceedings{panchenko2016taxi,
	title        = {A Taxonomy Induction Method based on Lexico-Syntactic Patterns, Substrings and Focused Crawling},
	author       = {Panchenko, Alexander  and Faralli, Stefano  and Ruppert, Eugen  and Remus, Steffen  and Naets, Hubert  and Fairon, C{\'e}drick  and Ponzetto, Simone Paolo  and Biemann, Chris},
	year         = 2016,
	booktitle    = {Proceedings of SemEval},
	pages        = {1320--1327},
}

@article{lipscomb2000medical,
	title        = {Medical subject headings (MeSH)},
	author       = {Lipscomb, Carolyn E},
	year         = 2000,
	journal      = {Bulletin of the Medical Library Association},
	pages        = 265
}

@inproceedings{luo2020alicoco,
	title        = {AliCoCo: Alibaba e-commerce cognitive concept net},
	author       = {Luo, Xusheng and Liu, Luxin and Yang, Yonghua and Bo, Le and Cao, Yuanpeng and Wu, Jinghang and Li, Qiang and Yang, Keping and Zhu, Kenny Q},
	year         = 2020,
	booktitle    = {Proceedings of SIGMOD},
	pages        = {313--327}
}

@inproceedings{mao2020octet,
	title        = {Octet: Online catalog taxonomy enrichment with self-supervision},
	author       = {Mao, Yuning and Zhao, Tong and Kan, Andrey and Zhang, Chenwei and Dong, Xin Luna and Faloutsos, Christos and Han, Jiawei},
	year         = 2020,
	booktitle    = {Proceedings of KDD},
	pages        = {2247--2257}
}

@article{vilnis2014word,
  title={Word representations via gaussian embedding},
  author={Vilnis, Luke and McCallum, Andrew},
  journal={arXiv preprint arXiv:1412.6623},
  year={2014}
}

@article{athiwaratkun2017multimodal,
  title={Multimodal word distributions},
  author={Athiwaratkun, Ben and Wilson, Andrew Gordon},
  journal={arXiv preprint arXiv:1704.08424},
  year={2017}
}

@inproceedings{athiwaratkun-etal-2018-probabilistic,
    title = "Probabilistic {F}ast{T}ext for Multi-Sense Word Embeddings",
    author = "Athiwaratkun, Ben  and
      Wilson, Andrew  and
      Anandkumar, Anima",
    booktitle = "Proceedings of ACL",
    year = "2018",
    pages = "1--11",
    publisher = "",
}

@inproceedings{taxbox,
    title = "Insert or Attach: Taxonomy Completion via Box Embedding",
    author = "Xue, Wei  and
      Shen, Yongliang  and
      Ren, Wenqi  and
      Guo, Jietian  and
      Pu, Shiliang  and
      Lu, Weiming",
    booktitle = "Proceedings of ACL",
    year = "2024",
    pages = "3851--3863",
}

@inproceedings{kg2e,
author = {He, Shizhu and Liu, Kang and Ji, Guoliang and Zhao, Jun},
title = {Learning to Represent Knowledge Graphs with Gaussian Embedding},
year = {2015},
isbn = {9781450337946},
booktitle = {Proceedings of CIKM},
pages = {623–632},
numpages = {10},
}

@article{Jebara,
author = {Jebara, Tony and Kondor, Risi and Howard, Andrew},
title = {Probability Product Kernels},
year = {2004},
issue_date = {12/1/2004},
volume = {5},
issn = {1532-4435},
journal = {J. Mach. Learn. Res.},
pages = {819–844},
numpages = {26}
}

@article{lecun2006tutorial,
  title={A tutorial on energy-based learning},
  author={LeCun, Yann and Chopra, Sumit and Hadsell, Raia and Ranzato, M and Huang, Fujie and others},
  journal={Predicting structured data},
  volume={1},
  number={0},
  year={2006}
}

@inproceedings{bansal-etal-2014-structured,
    title = "Structured Learning for Taxonomy Induction with Belief Propagation",
    author = "Bansal, Mohit  and
      Burkett, David  and
      de Melo, Gerard  and
      Klein, Dan",
    booktitle = "Proceedings of ACL",
    year = "2014",
    pages = "1041--1051"
}

@inproceedings{xu-etal-2023-tacoprompt,
    title = "{T}aco{P}rompt: A Collaborative Multi-Task Prompt Learning Method for Self-Supervised Taxonomy Completion",
    author = "Xu, Hongyuan  and
      Liu, Ciyi  and
      Niu, Yuhang  and
      Chen, Yunong  and
      Cai, Xiangrui  and
      Wen, Yanlong  and
      Yuan, Xiaojie",
    booktitle = "Proceedings of EMNLP",
    year = "2023",
    pages = "15804--15817",
}

@inproceedings{devlin2018bert,
  title={Bert: Pre-training of deep bidirectional transformers for language understanding},
  author={Kenton, Jacob Devlin Ming-Wei Chang and Toutanova, Lee Kristina},
  booktitle={Proceedings of naacL-HLT},
  volume={1},
  pages={2},
  year={2019},
  organization={Minneapolis, Minnesota}
}

@article{mahabal2023producing,
      title          = {Producing Usable Taxonomies Cheaply and Rapidly at Pinterest Using Discovered Dynamic $\mu$-Topics},
      author         = {A. Mahabal and Jiyun Luo and Rui Huang and Michael Ellsworth and Rui Li},
      year           = {2023},
      journal        = {ArXiv},
      volume         = {abs/2301.12520},
}

@inproceedings{jiang2023single,
      title          = {A Single Vector Is Not Enough: Taxonomy Expansion via Box Embeddings},
      author         = {Jiang, Song and Yao, Qiyue and Wang, Qifan and Sun, Yizhou},
      year           = {2023},
      pages          = {2467–2476},
      booktitle      = {WWW},
      isbn           = {9781450394161},
      numpages       = {10},
}

@inproceedings{snow2004learning,
      title          = {Learning Syntactic Patterns for Automatic Hypernym Discovery},
      author         = {Snow, Rion and Jurafsky, Daniel and Ng, Andrew},
      year           = {2004},
      volume         = {17},
      booktitle      = {NeurIPS},
}

@inproceedings{bordea2015semeval,
  title={Semeval-2015 task 17: Taxonomy extraction evaluation (texeval)},
  author={Bordea, Georgeta and Buitelaar, Paul and Faralli, Stefano and Navigli, Roberto and others},
  booktitle={Proceedings of {S}em{E}val},
  pages={902--910},
  year={2015},
}

@inproceedings{mishra2025rank,
    title = "Rank, Chunk and Expand: Lineage-Oriented Reasoning for Taxonomy Expansion",
    author = "Mishra, Sahil  and
      Arjun, Kumar  and
      Chakraborty, Tanmoy",
    booktitle = "Findings of ACL",
    year = "2025",
    pages = "12935--12953",
}
\appendix
\section*{Appendix}

\section{Implementation Details}
\label{app:implementation}
\modelname is implemented using PyTorch with the baselines taken from the respective repositories of their original authors. All training and inference tasks were conducted on a 48GB A6000 GPU. We adopt \texttt{bert-base-uncased} as the default encoder. The projection head comprises two 2-layer MLPs (hidden size 64) with a dropout rate of 0.2. Optimization uses AdamW with learning rates $9\times10^{-5}$ for BERT fine-tuning and $1\times10^{-3}$ for the projection layers. We train with a batch size of 128 for up to 125 epochs. The number of hard negatives per query is 50 for \textsc{Science}, 50 for \textsc{Environment}, 10 for \textsc{WordNet}, 20 for \textsc{MeSH}, and 20 for \textsc{Food}. Unless noted, loss weights are 0.45 for $\mathcal{L}_{\text{asym}}$, 0.45 for $\mathcal{L}_{\text{sym}}$, and 0.10 for the volume regularizer. For all baseline comparisons, we set $\lambda=0.3$ and $C$=1.5.

\section{Benchmark Datasets}
\label{app:data}
As discussed in Section \ref{subsubsec:data}, we evaluate \modelname\ on five public taxonomies, namely (i) Environment (ENV), (ii) Science (SCI), (iii) Food, (iv) WordNet, and (v) MeSH, summarized in Table \ref{table:dataset}. The dataset statistics are discussed in Table \ref{table:dataset}. ENV and SCI are small taxonomies that approximate hand-curated taxonomies by experts, WordNet is a medium-scale and domain-mixed taxonomy, reflecting noisier, application-agnostic settings in real-world, while Food and MeSH are large, multi-parent taxonomies that stress-test scalability and closely mirror real-world multi-parent settings where structures deviate from strict trees. Each dataset provides surface names and definitions for concepts. For evaluation, we build a seed taxonomy and hold out a set of query nodes by removing 20\% of leaf nodes while ensuring each query's gold parent remains in the seed. The remaining seed nodes supply self-supervision during training and serve as candidate anchors at inference.

\begin{table}[!t]
\centering
\caption{Statistics of benchmark datasets. ${|\mathcal{N}^0|}$ and ${|\mathcal{E}^0|}$ denote the number of nodes and edges in the seed taxonomy, while ${|D|}$ is the taxonomy depth. For WordNet, values are averaged across 114 sub-taxonomies.}
\begin{tabular}{lrrr}
\toprule
\textbf{Dataset} & \boldmath$|\mathcal{N}^0|$ & \boldmath$|\mathcal{E}^0|$ & \boldmath$|D|$ \\
\cmidrule(lr){1-4}
\multicolumn{4}{c}{\textbf{Single-parent taxonomies}} \\
\cmidrule(lr){1-4}
SemEval-Env      & 261  & 261   & 6  \\
SemEval-Sci      & 429  & 452   & 8  \\
WordNet      & 20.5 & 19.5  & 3  \\
\addlinespace[2pt]
\cmidrule(lr){1-4}
\multicolumn{4}{c}{\textbf{Multi-parent taxonomies}} \\
\cmidrule(lr){1-4}
SemEval-Food     & 1486 & 1533  & 8  \\
MeSH             & 9710 & 10498 & 12 \\
\bottomrule
\end{tabular}
\label{table:dataset}
\end{table}

\begin{table}[!t]
\centering
\caption{Impact of the asymmetric optimization ($\mathcal{L}_\text{asym}$) on ENV, and WordNet benchmarks. ``$\downarrow$'' indicates that lower values denote better performance. `W/O' means \textit{without}.}
\label{table:no_kl_abl}
\begin{tabular}{@{}c|rrr|rrr@{}}
\toprule
\addlinespace[2pt]
\multirow{2}{*}{\textbf{Method}} &
\multicolumn{3}{c|}{\textbf{Environment}} &
\multicolumn{3}{c}{\textbf{WordNet}} \\
\cmidrule(lr){2-4}\cmidrule(lr){5-7}
& \textbf{H@1} & \textbf{MR$\downarrow$} & \textbf{MRR}
& \textbf{H@1} & \textbf{MR$\downarrow$} & \textbf{MRR} \\
\midrule
\textbf{W/O $\mathcal{L}_\text{asym}$} & 44.23 & 11.51 & 53.55 & 34.58 & 47.53 & 47.48 \\
\rowcolor{gray!20}\textbf{$\mathcal{L}_\text{asym}$} & 52.20 & \,8.98 & 58.36 & 37.40 & 41.00 & 52.40 \\
\midrule
$\uparrow\%$ & +18.02 & +21.98 & +8.98 & +8.16 & +13.74 & +10.36 \\
\bottomrule
\end{tabular}
\end{table}

\section{Baselines}
\label{app:baselines}

We compare \modelname\ with a diverse set of the following baselines outlined in Section~\ref{subsubsec:baseline},

\begin{itemize}[leftmargin=*]
  \item \textbf{BERT+MLP} \cite{devlin2018bert} encodes term surface forms with \textsc{BERT} and feeds the resulting vectors to a multi-layer perceptron to classify hypernym relations.  
  \item \textbf{TaxoExpan} \cite{shen2020taxoexpan} represents an anchor node by encoding its ego network with a graph neural network and scores parent–child pairs through a log-bilinear feed-forward layer.
  \item \textbf{Arborist} \cite{manzoor2020expanding} models heterogeneous edge semantics and trains with a large-margin ranking loss whose margin adapts dynamically.
  \item \textbf{BoxTaxo} \cite{jiang2023single} learns box embeddings and evaluates parent candidates with geometric and probabilistic losses derived from hyper-rectangle volumes.
  \item \textbf{TMN} \cite{zhang2021taxonomy} method employs subtasks, namely attaching the query to the parent and the child to the query, as auxiliary supervision signals for concept representation learning.
  \item \textbf{STEAM} \cite{yu_steam_2020} ensembles graph, contextual, and lexical-syntactic features in a multi-view co-training framework to score hypernymy links.  
  \item \textbf{TaxoEnrich} \cite{jiang2022taxoenrich} utilizes structural information through taxonomy-contextualized embeddings, enhancing position representations with a query-aware sibling aggregator.
\end{itemize}

TaxoEnrich and TMN are designed for taxonomy completion. They insert a query $q$ between a parent–child pair and therefore score triplets $f(p,c,q)$. Our setting is taxonomy expansion, where q is attached as a leaf under a parent and no child $c$ is available. For a fair comparison, we follow \cite{wang_qen_2022} and adapt these baselines by instantiating $c$ with a dummy placeholder (e.g., a blank/sentinel token). This converts their triplet scorer to an effective leaf-attachment scheme while preserving their original scoring function.

\section{Evaluation Metrics}
\label{app:metrics}
Given the query set \(\mathcal{C}\), let the predictions made by the model be \(\left\{\hat{y}_1, \hat{y}_2, \ldots, \hat{y}_{|\mathcal{C}|}\right\}\) and the ground-truth positions be \(\left\{y_1, y_2, \cdots, y_{|\mathcal{C}|}\right\}\). Following \citet{liu2021temp, manzoor2020expanding, vedula_enriching_2018, jiang2022taxoenrich} and \citet{yu_steam_2020}, we use the following metrics to evaluate the performance of the competing models.

\begin{itemize}[leftmargin=*]
\item \textbf{Hit@k}: It measures the fraction of instances where the correct label appears among the top-$k$ predictions.  
It quantifies how often the model successfully ranks the true class within its top-$k$ guesses.  
Formally, a hit is counted if $y_i \in \hat{Y}_i^{(k)}$ for the $i$-th instance.  
Higher Hit@k indicates better retrieval or ranking performance.
\begin{equation}
\text{Hit@}k = \frac{1}{|\mathcal{C}|} \sum_{i=1}^{|\mathcal{C}|} \mathbb{I}(y_i \in \hat{Y}_i^{(k)}),
\end{equation}
where $\hat{Y}_i^{(k)}$ denotes the set of top-$k$ predicted labels for the $i$-th instance, and $\mathbb{I}(\cdot)$ is the indicator function that equals 1 if the ground-truth label $y_i$ is present in the top-$k$ predictions, and 0 otherwise.
\item \textbf{Recall@k}: It measures the proportion of ground-truth labels that are correctly retrieved  
within the top-$k$ predictions for each instance. It reflects the model's ability to capture  
relevant labels among its top-ranked outputs. Formally, it is defined as:
\begin{equation}
\text{Recall@}k = \frac{1}{|\mathcal{C}|} \sum_{i=1}^{|\mathcal{C}|} 
\frac{|\{ y \in \mathcal{Y}_i : y \in \hat{Y}_i^{(k)} \}|}{|\mathcal{Y}_i|},
\end{equation}
where $\mathcal{Y}_i$ is the set of true labels and $\hat{Y}_i^{(k)}$ the top-$k$ predicted labels. Note that for single parent taxonomies, where each instance has exactly one ground-truth label  
($|\mathcal{Y}_i| = 1$), Recall@k becomes equivalent to Hit@k.
\item \textbf{Mean Rank (MR)}: It measures the average position (rank) of the correct label among all predicted scores. It evaluates how highly the model ranks the ground-truth label across all instances. A lower mean rank indicates better performance, as correct labels appear earlier in the ranking. Formally, it is defined as:
\begin{equation}
\text{MR} = \frac{1}{|\mathcal{C}|} \sum_{i=1}^{|\mathcal{C}|} \text{rank}(y_i),
\end{equation}
where $\text{rank}(y_i)$ denotes the position of the true label $y_i$ in the sorted list of  
predicted scores for the $i$-th instance (rank 1 being the best).
\item \textbf{Mean Reciprocal Rank (MRR)}: It evaluates the average inverse rank of the correct label across all instances. It emphasizes not only whether the true label appears in the predictions  
but also how highly it is ranked. Higher MRR values indicate that correct labels are placed  
closer to the top of the ranked list. Formally, it is defined as:
\begin{equation}
\text{MRR} = \frac{1}{|\mathcal{C}|} \sum_{i=1}^{|\mathcal{C}|} \frac{1}{\text{rank}(y_i)},
\end{equation}
where $\text{rank}(y_i)$ denotes the position of the ground-truth label $y_i$ in the model's  
predicted ranking for the $i$-th instance.
\item \textbf{Wu \& Palmer (Wu\&P) Similarity}: It measures the semantic similarity between two concepts  
based on their depth in a taxonomy and the depth of their lowest common ancestor (LCA).  
It quantifies the degree of relatedness between two nodes within a hierarchical structure. For  
single-parent taxonomies, where each node has a unique ancestor path, the measure  
is defined as:
\begin{equation}
\text{Wu\&P}(c_1, c_2) = \frac{2 \times \text{depth}(\text{LCA}(c_1, c_2))}{\text{depth}(c_1) + \text{depth}(c_2)},
\end{equation}
where $\text{depth}(c)$ denotes the distance of the concept $c$ from the root, and  
$\text{LCA}(c_1, c_2)$ is their lowest common ancestor.

\end{itemize}

\begin{table}[t]
\centering
\caption{Impact of the symmetric optimization ($\mathcal{L}_\text{sym}$) on ENV, and WordNet benchmarks. ``$\downarrow$'' indicates that lower values denote better performance. `W/O' means \textit{without}.}
\label{table:no_bc_abl}
\begin{tabular}{@{}c|rrr|rrr@{}}
\toprule
\addlinespace[2pt]
\multirow{2}{*}{\textbf{Method}} &
\multicolumn{3}{c|}{\textbf{Environment}} &
\multicolumn{3}{c}{\textbf{WordNet}} \\
\cmidrule(lr){2-4}\cmidrule(lr){5-7}
& \textbf{H@1} & \textbf{MR$\downarrow$} & \textbf{MRR}
& \textbf{H@1} & \textbf{MR$\downarrow$} & \textbf{MRR} \\
\midrule
\textbf{W/O $\mathcal{L}_\text{sym}$} & 13.46 & \,62.01 & 18.06 & \,8.87 & 228.64 & \,8.71 \\
\rowcolor{gray!20}\textbf{$\mathcal{L}_\text{sym}$}    & 50.91 & \,8.89  & 61.20 & 36.84 & \,45.17 & 48.96 \\
\midrule
$\uparrow\%$   & +278.2 & +85.7 & +239.0 & +315.3 & +80.2 & +462.0 \\
\bottomrule
\end{tabular}
\end{table}

\begin{table}[t]
\centering
\caption{Effect of the volume regularization ($\mathcal{L}_\text{reg}$ and $\mathcal{L}_\text{clip}$) on ENV, and WordNet benchmarks. ``$\downarrow$'' indicates that lower values denote better performance. `W/O' means \textit{without}.}
\label{table:vol_reg_abl}
\begin{tabular}{@{}c|rrr|rrr@{}}
\toprule
\addlinespace[2pt]
\multirow{2}{*}{\textbf{Method}} &
\multicolumn{3}{c|}{\textbf{Environment}} &
\multicolumn{3}{c}{\textbf{WordNet}} \\
\cmidrule(lr){2-4}\cmidrule(lr){5-7}
& \textbf{H@1} & \textbf{MR$\downarrow$} & \textbf{MRR}
& \textbf{H@1} & \textbf{MR$\downarrow$} & \textbf{MRR} \\
\midrule
\textbf{W/O} & 38.46 & 13.25 & 49.73 & 32.24 & 53.48 & 46.27 \\
\rowcolor{gray!20}\textbf{Reg} & 52.20 & \,8.98 & 58.36 & 37.40 & 41.00 & 52.40 \\
\midrule
\textbf{$\uparrow\%$} & +35.73 & +32.23 & +17.35 & +16.00 & +23.34 & +13.25 \\
\bottomrule
\end{tabular}
\end{table}

\begin{table}[t]
\centering
\caption{Effect of $\mathcal{L}_\text{diverge}$ on ENV, and WordNet benchmarks. ``$\downarrow$'' indicates that lower values denote better performance. `W/O' means \textit{without}.}
\label{table:diverge_abl}
\begin{tabular}{@{}c|rrr|rrr@{}}
\toprule
\addlinespace[2pt]
\multirow{2}{*}{\textbf{Method}} &
\multicolumn{3}{c|}{\textbf{Environment}} &
\multicolumn{3}{c}{\textbf{WordNet}} \\
\cmidrule(lr){2-4}\cmidrule(lr){5-7}
& \textbf{H@1} & \textbf{MR$\downarrow$} & \textbf{MRR}
& \textbf{H@1} & \textbf{MR$\downarrow$} & \textbf{MRR} \\
\midrule
\textbf{W/O $\mathcal{L}_\text{diverge}$} & 44.23 & 10.40 & 59.27 & 35.32 & 61.23 & 39.12 \\
\rowcolor{gray!20}\textbf{$\mathcal{L}_\text{diverge}$} & 52.20 & \,8.98 & 58.36 & 37.40 & 41.00 & 52.40 \\
\midrule
\textbf{$\uparrow\%$} & +18.02 & +13.65 & $-1.54$ & +5.89 & +33.04 & +33.95 \\
\bottomrule
\end{tabular}
\end{table}

\section{Ablations}
\label{app:ablations}
As a complement to Section \ref{subsec:ablations}, we report additional ablations on ENV and WordNet in Tables \ref{table:no_kl_abl}, \ref{table:no_bc_abl}, \ref{table:vol_reg_abl}, and \ref{table:diverge_abl}. The same pattern is evident in both benchmarks. Removing the asymmetric containment loss compromises hierarchical directionality, causing queries to drift to nearby but structurally incorrect anchors. Dropping the symmetric overlap loss weakens semantic cohesion and produces brittle neighborhoods, which is the most damaging setting and leads to large drops in ranking quality. Eliminating volume regularization allows covariances to collapse or explode, breaking the parent–child size relation and degrading all metrics. Adding the reverse-KL coverage term improves results relative to using alignment alone because it keeps parents broader than their children and stabilizes the training process. Taken together, these results demonstrate that overlap, directional containment, coverage control, and volume regularization work in tandem to produce calibrated Gaussian boxes and reliable attachments, and that removing any single component consistently degrades performance.

\begin{figure}
    \centering
    \includegraphics[width=1.0\linewidth]{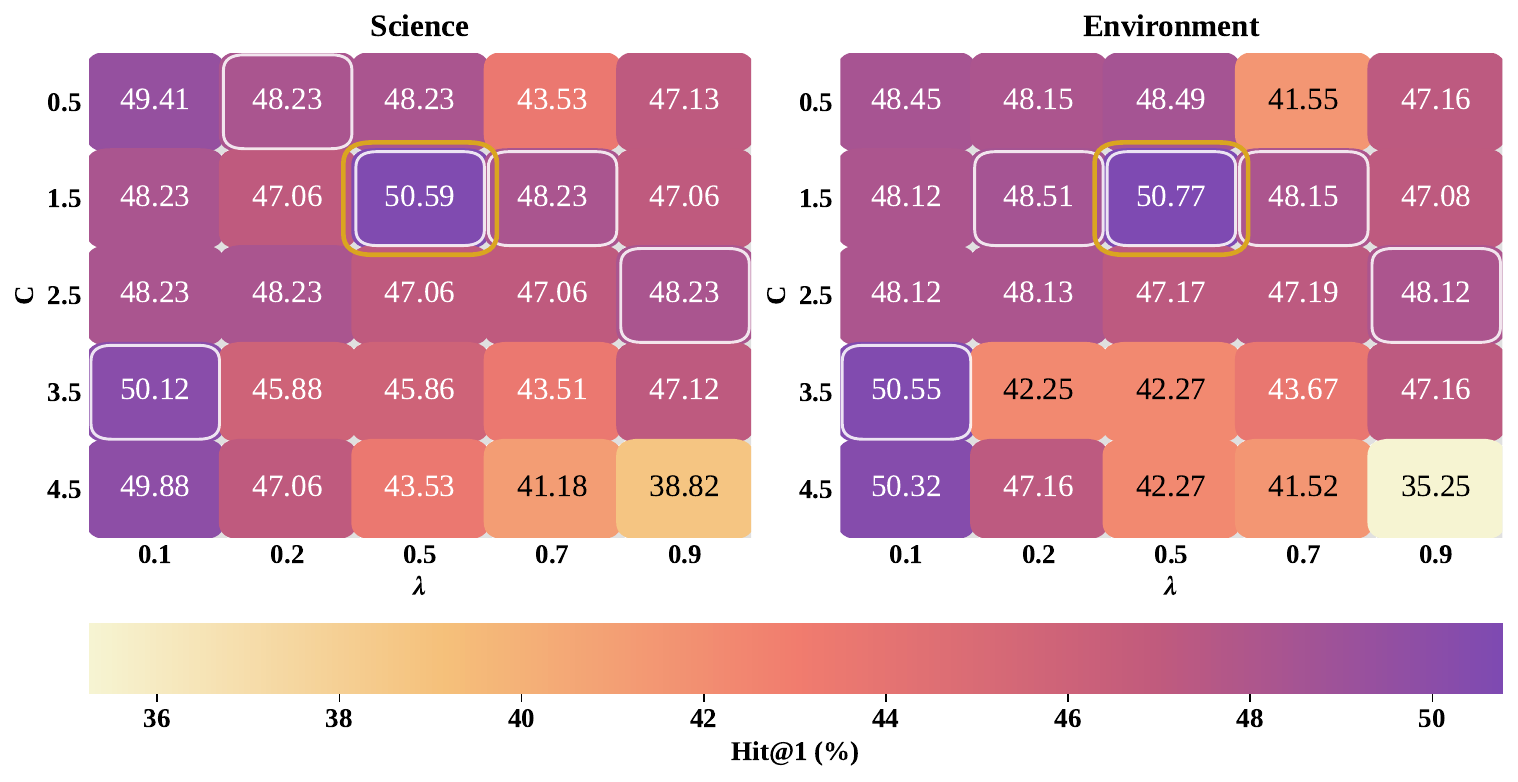}
    \caption{Heatmap of sensitivity of the $\mathcal{L}_\text{diverge}$ hyperparameters on \textbf{Hit@1} for \textbf{Science} and \textbf{Environment}. We sweep the scale $C$ (vertical axis) and the weight $\lambda$ (horizontal axis).}
    \Description{Heatmap showing the effect of scale C and weight lambda on Hit@1 for Science and Environment.}
    \label{fig:lambda_c_ablation}
\end{figure}

\begin{table}[t]
\centering
\caption{Qualitative examples of \modelname\ on the MeSH and Food benchmarks (Multi-Parent). For each benchmark, we show one correct and one incorrect attachment. Correct predictions are marked with \cmark while incorrect with \xmark.}
\label{table:multi_p_examples}
\setlength{\tabcolsep}{2pt}
\renewcommand{\arraystretch}{1.2}
\scalebox{0.82}{
\begin{tabular}{@{}p{2.8cm}p{3.6cm}p{3.6cm}@{}}
\toprule
\textbf{Query} & \textbf{Anchor (Gold)} & \textbf{Top Predictions} \\
\midrule
\rowcolor{gray!20}\multicolumn{3}{@{}c}{\textbf{MeSH}}\\
Copper & Transition elements, Heavy metals &
  \begin{tabular}[t]{@{}l@{}}
  Transition elements \cmark\\
  Tin \xmark\\
  Heavy metals \cmark
  \end{tabular} \\
\rowcolor{gray!10}
Mitochondrial membrane transport proteins &
Mitochondrial proteins,   Membrane transport proteins &
  \begin{tabular}[t]{@{}l@{}}
  Cation transport proteins \xmark\\
  Electron transport chain-\\ complex proteins \xmark\\
  NADH dehydrogenase \xmark
  \end{tabular} \\
  Translational protein modification &
Gene expression regulation, Protein biosynthesis &
  \begin{tabular}[t]{@{}l@{}}
  Peptide biosynthesis\xmark\\
  Translational peptide chain-\\ termination \xmark\\
  Protein biosynthesis\cmark
  \end{tabular} \\
\midrule
\rowcolor{gray!20}\multicolumn{3}{@{}c}{\textbf{Food}}\\
Frozen orange juice & Orange juice, Concentrate &
  \begin{tabular}[t]{@{}l@{}}
  Fruit juice \xmark\\
  Orange juice \cmark\\
  Grapefruit juice \xmark
  \end{tabular} \\
\rowcolor{gray!10}
Mocha & Coffee, Flavorer &
  \begin{tabular}[t]{@{}l@{}}
  Dessert \xmark\\
  Charlotte \xmark\\
  Course \xmark
  \end{tabular} \\
  Onion bread & Bread &
  \begin{tabular}[t]{@{}l@{}}
  Rye bread \xmark\\
  Bread \cmark\\
  Black bread \xmark
  \end{tabular} \\
\bottomrule
\end{tabular}}

\end{table}

\section{Scaling and Dynamic Margin factor}
\label{app:scale}
To complement the ablations in Section~\ref{subsec:ablations}, we tune the two hyperparameters in the divergence term $\mathcal{L}_{\text{diverge}}$, namely the scale $C$ and the weight $\lambda$. The scale $C$ controls how much broader a parent should be than its child and sets the required information gap. The weight $\lambda$ sets the overall influence of $\mathcal{L}_{\text{diverge}}$ in the objective. Fig. \ref{fig:lambda_c_ablation} reports results for Science and Environment. A moderate $C$ improves performance by enforcing a meaningful coverage gap. A very large $C$ becomes too restrictive, degrading parent-child alignment. The weight $\lambda$ shows a similar trade-off. Very small values are underregularized, while very large values push the model toward a rigid geometric pattern that shrinks child variances and inflates parent variances. This reduces expressiveness and weakens semantic alignment. In practice, we find $C \approx 1.5$ and $\lambda \approx 0.5$ to be a good operating point. These settings preserve the intuition that parents should be wider than children while keeping the embedding space well-calibrated.

\section{Case Study}
\label{app:case_study}
As discussed in Section \ref{subsec:case}, we include additional multi-parent case studies for the MeSH and SemEval-Food benchmarks as shown in Table \ref{table:multi_p_examples}. On \textbf{MeSH}, \emph{Copper} is correctly linked to both \emph{Transition elements} and \emph{Heavy metals}, indicating that the model can allocate probability mass to cover more than one valid parent, with Tin also ranked in top-$3$ because copper and tin are closely related metals that are often co-mentioned (e.g., bronze is an alloy of copper and tin), their definitions and usage strongly overlap, yielding a high symmetric-overlap score. In contrast, \emph{Mitochondrial membrane transport proteins} is misattached to nearby transport categories because the query shares strong lexical and semantic cues such as \emph{mitochondrial}, \emph{membrane}, \emph{transport}, and \emph{proteins}, with several sibling classes whose MeSH definitions describe closely related processes, which boosts symmetric similarity across candidates and, under a shared ``transport'' ancestor, weakens the KL-based preference for the correct parent. On \textbf{SemEval-Food}, \emph{Frozen orange juice} returns several reasonable parents (\emph{Fruit juice}, \emph{Orange juice}). This shows the model can capture both broad categories and specific products. It also ranks \emph{Fruit juice} above the gold label \emph{Concentrate}, so the top result looks \textit{wrong} even though Fruit juice is arguably a better parent. In contrast, \emph{Mocha} leans toward dessert-related parents, which likely comes from ambiguity between drinks and sweets. In general, the model does well on multi-parent cases when the shared meaning across parents is clear. Mistakes happen when sibling categories are very similar or the query is ambiguous. Better definitions, harder negatives from the same family, and tighter uncertainty control may further improve these cases.


\end{document}